\documentclass{article}

\usepackage{microtype}
\usepackage{graphicx}
\usepackage{subfigure}
\usepackage{subcaption}
\usepackage{booktabs} 
\usepackage{threeparttable} 

\usepackage{hyperref}

\usepackage[accepted]{icml2024}

\usepackage{amsmath}
\usepackage{amssymb}
\usepackage{mathtools}
\usepackage{amsthm}
\usepackage{xcolor}
\usepackage{color,colortbl}
\usepackage{multirow}
\usepackage{siunitx}
\usepackage{tabularx}
\usepackage{booktabs}
\usepackage{makecell}
\usepackage{soul}
\usepackage{footnote}
\makesavenoteenv{tabular}
\makesavenoteenv{table}
\usepackage{pifont}
\usepackage[capitalize,noabbrev]{cleveref}

\definecolor{mycolor1}{rgb}{0.9569, 0.9569, 0.9569}
\definecolor{mycolor2}{rgb}{0.8980, 0.9961, 0.8275}
\definecolor{mycolor3}{rgb}{0.5529, 0.8824, 0.5529}
\definecolor{mycolor4}{rgb}{0.4745, 0.7686, 0.4353}

\theoremstyle{plain}
\newtheorem{theorem}{Theorem}[section]

\theoremstyle{definition}
\newtheorem{definition}[theorem]{Definition}

\theoremstyle{remark}

\usepackage[textsize=tiny]{todonotes}

\icmltitlerunning{Towards Unified Alignment Between Agents, Humans, and Environment}

\begin{document}

\twocolumn[
\icmltitle{Towards Unified Alignment Between Agents, Humans, and Environment}



\icmlsetsymbol{equal}{*}

\begin{icmlauthorlist}
\icmlauthor{Zonghan Yang}{thu,equal}
\icmlauthor{An Liu}{thu,equal}
\icmlauthor{Zijun Liu}{thu,equal}
\icmlauthor{Kaiming Liu}{thu}
\icmlauthor{Fangzhou Xiong}{thu}
\icmlauthor{Yile Wang}{thu}
\icmlauthor{Zeyuan Yang}{thu}
\icmlauthor{Qingyuan Hu}{thu}
\icmlauthor{Xinrui Chen}{thu}
\icmlauthor{Zhenhe Zhang}{thu}
\icmlauthor{Fuwen Luo}{thu}
\icmlauthor{Zhicheng Guo}{thu}
\icmlauthor{Peng Li}{air}
\icmlauthor{Yang Liu}{thu,air}

\end{icmlauthorlist}

\icmlaffiliation{thu}{Department of Computer Science and Technology, Tsinghua University, Beijing, China}
\icmlaffiliation{air}{Institute for AI Industry Research (AIR), Tsinghua University, Beijing, China}

\icmlcorrespondingauthor{Peng Li}{lipeng@air.tsinghua.edu.cn}
\icmlcorrespondingauthor{Yang Liu}{liuyang2011@tsinghua.edu.cn}

\icmlkeywords{Machine Learning, ICML}

\vskip 0.3in
]



\printAffiliationsAndNotice{\icmlEqualContribution} 

\begin{abstract}
The rapid progress of foundation models has led to the prosperity of autonomous agents, which leverage the universal capabilities of foundation models to conduct reasoning, decision-making, and environmental interaction. However, the efficacy of agents remains limited when operating in intricate, realistic environments. In this work, we introduce the principles of \textbf{U}nified \textbf{A}lignment for \textbf{A}gents (\textbf{UA}$^2$), which advocate for the simultaneous alignment of agents with human intentions, environmental dynamics, and self-constraints such as the limitation of monetary budgets. From the perspective of \textbf{UA}$^2$, we review the current agent research and highlight the neglected factors in existing agent benchmarks and method candidates. We also conduct proof-of-concept studies by introducing realistic features to WebShop~\cite{yao2022webshop}, including user profiles to demonstrate intentions, personalized reranking for complex environmental dynamics, and runtime cost statistics to reflect self-constraints. We then follow the principles of \textbf{UA}$^2$ to propose an initial design of our agent and benchmark its performance with several candidate baselines in the retrofitted WebShop. The extensive experimental results further prove the importance of the principles of \textbf{UA}$^2$. Our research sheds light on the next steps of autonomous agent research with improved general problem-solving abilities.

\end{abstract}

\section{Introduction}

Recent days have witnessed the rapid development of autonomous agents, which leverage the proficiency of Large Language Models (LLMs) or Large Multimodal Models (LMMs)~\cite{gpt4,touvron2023llama2,team2023gemini,jiang2024mixtral} to interact with environments for task execution. Several seminal works on foundation model agents have exhibited promising results in both digital and embodied scenarios, including but not limited to web task automation~\cite{deng2023mind2web,zhou2023webarena,zheng2023seeact}, open-ended world exploration~\cite{wang2023voyager,zhu2023ghost}, interactive coding~\cite{chen2023teaching,qian2023communicative,xu2023lemur}, and robotic tasks~\cite{saycan2022arxiv,generalpatternmachines2023,huang2023voxposer,ma2023eureka,wang2023robogen}. 

Aside from existing literature, the development of foundation model agents in realistic, complex scenarios is still in its infancy. While different agent benchmarks have been proposed~\cite{liu2023agentbench,mialon2023gaia,ma2024agentboard}, the methodologies of agents are still being proposed and evaluated in synthetic, simplified settings, which results in the bottlenecked performance of agents in real-world deployment when attempting to satisfy the expectations of humans with realistic demands~\cite{kinniment2023evaluating}. This leads to the question: \textit{What are the principles the agents should follow to improve their real-world capabilities?}

\begin{figure}[t]
    \centering
    \includegraphics[width=0.48\textwidth]{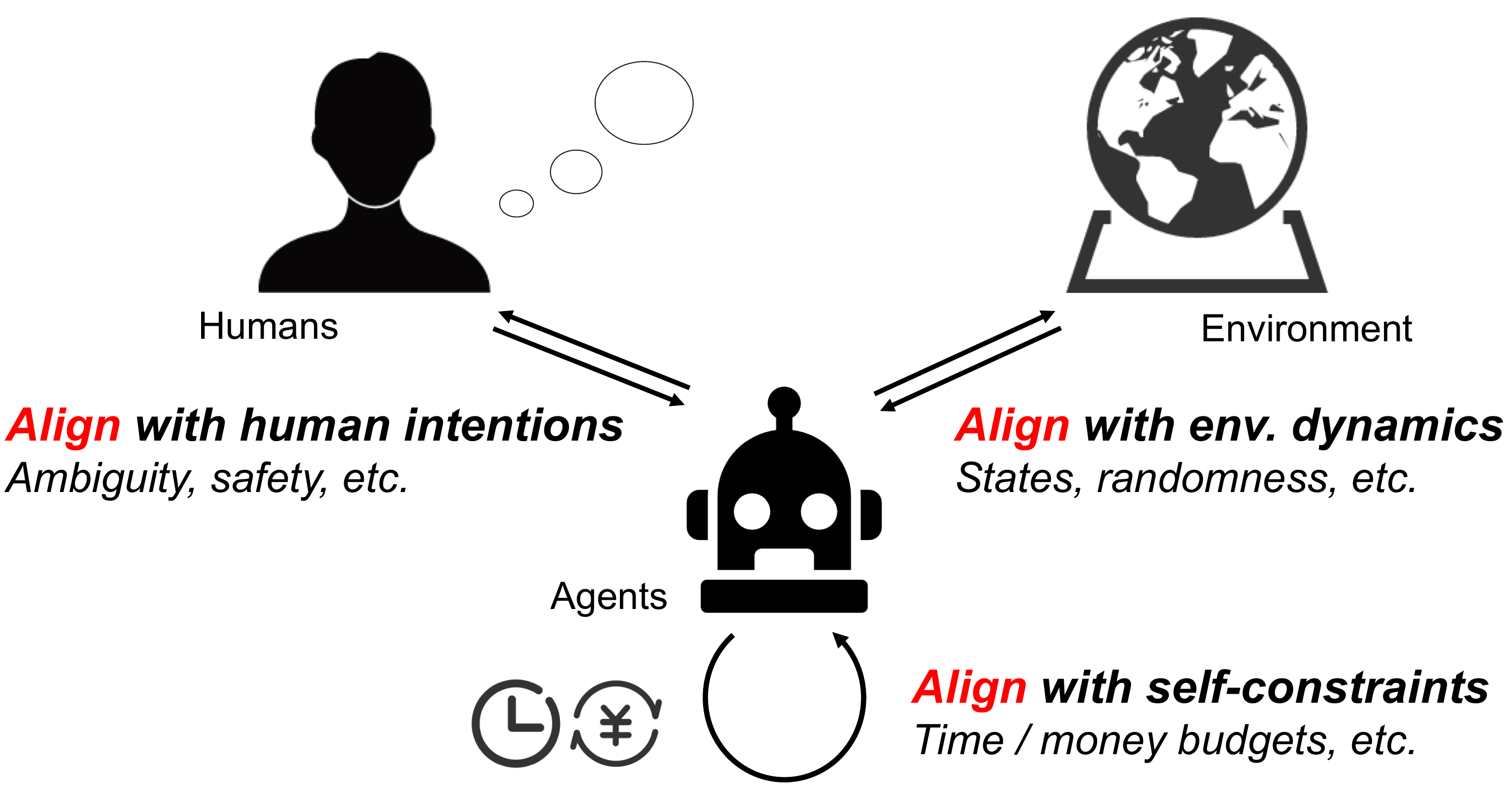}
    \caption{A working system of agents that consists of three roles: humans to be assisted, an environment to interact with, and the agents themselves. The principles of \textbf{U}nified \textbf{A}lignment for \textbf{A}gents (\textbf{UA}$^2$) suggest that the agents should align with the three roles in a unified manner by recognizing \textit{human intentions}, adapting to \textit{environmental dynamics}, and adhering to \textit{self-constraints}. }
    \label{fig-1}
    \vspace{-10pt}
\end{figure}

\begin{figure*}[ht]
    \centering
    \includegraphics[width=0.98\textwidth]{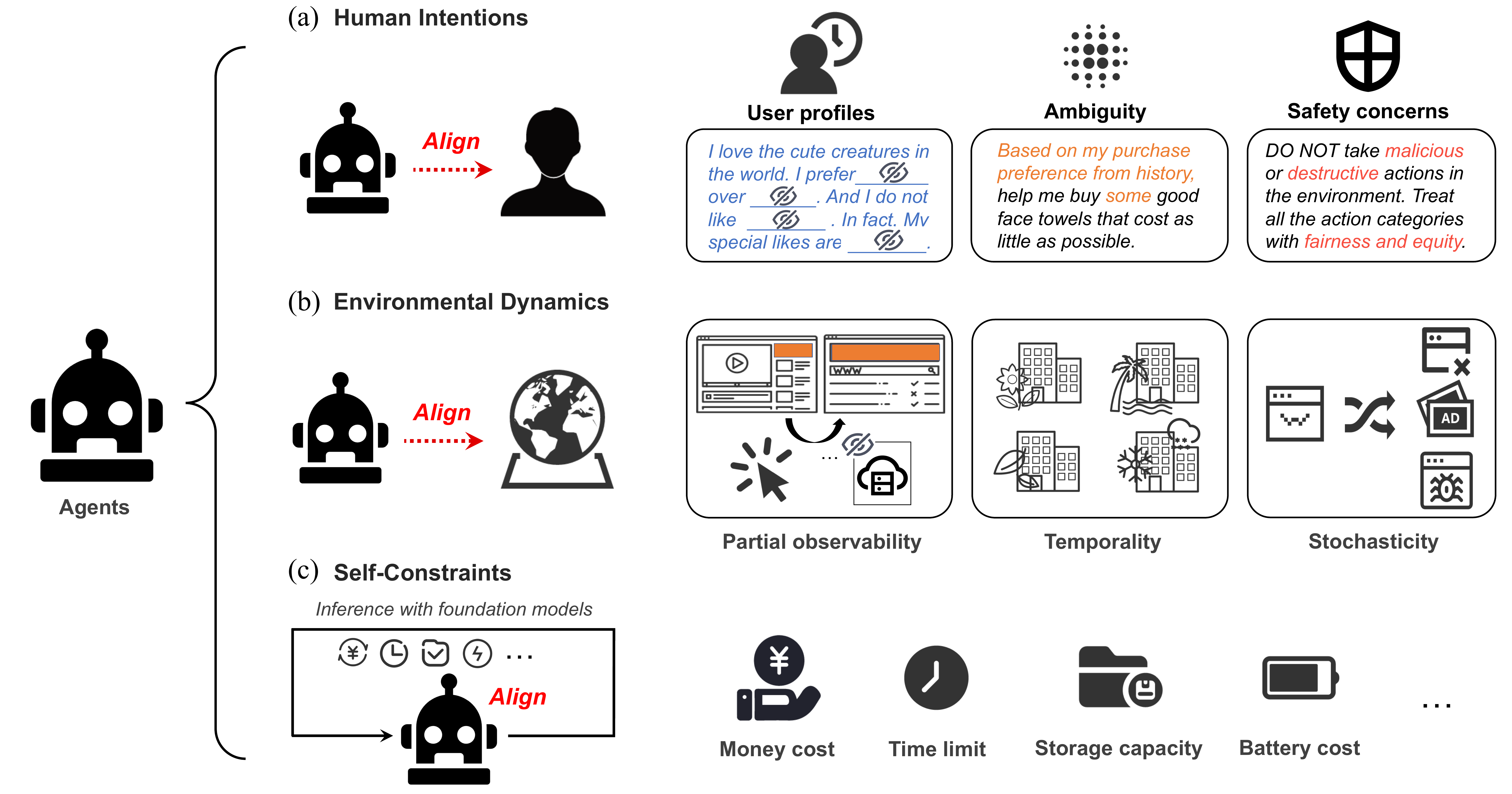}
    \caption{Illustrations of the principles of unified alignment with (a) \textit{human intentions}, (b) \textit{environmental dynamics}, and (c) \textit{self-constraints}. The principles of unified alignment for agents emerge from all the roles in an agent working system: agents, humans, and environment.}
    \label{fig-2}
\end{figure*}

\looseness=-1 To answer the question, we first take a systematic view of agents. We recognize the working system of agents as a composition of three roles: \textit{humans} that propose the goals to be assisted, an \textit{environment} that provides feedback for interaction, and foundation model \textit{agents} themselves to act in the environment to assist the human user. In complex scenarios, the intentions of humans can be ambiguous~\cite{tamkin2022task, li2023eliciting} or concerned with safety requirements~\cite{ruan2023identifying,yuan2024r}. Moreover, the underlying dynamics of the environment can be complicated to identify~\cite{lecun2022path,hu2023language}, and affected by temporality~\cite{zhou2024hazard} or stochasticity~\cite{wu2023smartplay}. Last but not least, the agents themselves can also be constrained by a certain amount of budgetary limits (\textit{e.g.}, monetary and time expenses) during operations, an aspect often overlooked in the existing agent research. While each of the aspects is noted by different aforementioned works, none of them emphasize the holistic comprehension of all the roles in the working system. 

In this work, we propose the principles of \textbf{U}nified \textbf{A}lignment for \textbf{A}gents (\textbf{UA}$^2$) by drawing connections with the alignment research in the sense of both LLMs and reinforcement learning literature~\cite{sutton2018reinforcement,ouyang2022training,bai2022constitutional,ji2023ai,burns2023weak}. The goal of \textbf{UA}$^2$ is to enhance the awareness of the foundation model agents to their working system, aligning with \textit{human intentions}, \textit{environmental dynamics}, and \textit{self-constraints} in a unified manner. From the perspective of \textbf{UA}$^2$, we review the existing research on agents and point out the neglected factors in the design of existing benchmarks and candidate methodologies of foundation model agents.

To further demonstrate the essence of \textbf{UA}$^2$, we conduct proof-of-concept studies by constructing an upgraded version of WebShop~\cite{yao2022webshop}. In the retrofitted WebShop, we add the design of the \textit{human intentions} of shoppers for agents to track and infer, the \textit{environmental dynamics} with personalized re-ranking algorithms that evolve with agent actions, and the \textit{self-constraints} by implementing a counter of monetary and temporal costs. On top of the retrofitted environment, we initiate an agent method guided by the principles of \textbf{UA}$^2$, and benchmark its performance as well as several other candidate agent baselines. The results reveal the suboptimality of the agent baselines that violate the principles of \textbf{UA}$^2$. The results further support our advocacy that the agents should achieve a unified alignment with humans, the environment, and the agents themselves. Our research sheds light on the future steps of autonomous agents, including synergizing agents with alignment techniques, constructing agent benchmarks and methods that follow the principles of \textbf{UA}$^2$, and envisioning self-evolving agents through lifelong interaction and continual alignment.

\section{Principles of Unified Alignment for Agents}

\subsection{Roles in a Working System of Agents}

We primarily discuss the three roles in an agent working system (Figure~\ref{fig-1}): agents, humans, and the environment. 

\textit{Agents} are the core component of the entire system. Agents are responsible for understanding human intentions and generating appropriate responses or actions to interact with the environment. Proficient agents should provide accurate, informative, and engaging interactions during task execution. 

\textit{Humans} are the main role to be assisted in the system. The tasks assigned by humans can be viewed as the initial inputs to the working system (especially to the agents), which reflects the underlying goals and human intentions.

\textit{Environment} refers to the situation where the agents operate. It encompasses the external factors and conditions that can influence the agents' behavior, performance, and interactions. The feedback from the environment affects the reasoning of the agents, as well as their following actions. 

Realistic working systems of agents are composed of diverse, ambiguous human intentions, changing environments with complex dynamics, as well as self-constraints over the agents themselves. This leads to the necessity of agents to operate towards the unified alignment with all the roles.

\subsection{Unified Alignment with All the Roles}

\begin{table*}[ht!]
\caption{Existing agent benchmarks from the perspective of alignment with \textit{human intentions}, \textit{environmental dynamics}, and \textit{self-constraints}. ``Temp.'' stands for temporality, and ``Stoch.'' stands for stochasticity. ``\#Actions'' means that the step counts / interaction turns in the environment will be considered as a metric. $\dagger$ WebShop is fully observable as long as the URL is covered in each observation.}
\label{table:agents}
\centering
{
\resizebox{1.0\linewidth}{!}{
\begin{threeparttable}
\begin{tabular}{@{}llcccccc@{}}

\toprule
\textbf{Type}&\textbf{Benchmarks}  & \textbf{Human Intentions} & \textbf{Environmental Dynamics} & \textbf{Self-Constraints} \\
\midrule

\multirow{5}*{Digital}
&Androidenv~\cite{toyama2021androidenv}   & None & Partial Obs. & None\\
&WebShop~\cite{yao2022webshop}    & None & Full Obs.$^{\dagger}$ & None \\
&Mind2Web~\cite{deng2023mind2web}   & None & Partial Obs. & None \\
&ToolBench~\cite{qin2023toolllm}        & None & Full Obs. \& Temp. \& Stoch. & None \\
&WebArena~\cite{zhou2023webarena}        & Fixed and Given &  Partial Obs.  & None \\
\midrule

\multirow{10}*{Embodied}
&VirtualHome~\cite{puig2018virtualhome}   &None& Partial Obs. &None \\
&BabyAI~\cite{chevalierboisvert2019babyai}&None& Partial Obs. &None\\
&ALFWorld~\cite{shridhar2020alfworld}    & None & Partial Obs. & None \\
&MineDojo~\cite{NEURIPS2022_74a67268}   & None & Partial Obs. \& Stoch.  & None\\
&ScienceWorld~\cite{wang2022scienceworld}&None& Partial Obs. &None\\
&Interactive Gibson~\cite{xia2020interactive} & None & Partial Obs. & \#Actions\\
&AGENT~\cite{shu2021agent} & None & Partial Obs. & \#Actions \\
&RFUniverse~\cite{fu2022rfuniverse} & Fixed and Given & Partial Obs. & \#Actions\\
&BEHAVIOR-1K~\cite{li2023behavior} & None & Full Obs. & \#Actions\\
&HAZARD~\cite{zhou2024hazard} & None & Partial Obs. \& Temp. & \#Actions \\
\midrule

\multirow{4}*{Mixed}
&MINT~\cite{wang2023mint} & None & Partial Obs. & \#Actions \\
&SmartPlay~\cite{wu2023smartplay} & None & Partial Obs. \& Stoch. & None \\
&AgentBench~\cite{liu2023agentbench}&None& Partial Obs. &None\\
&AgentBoard~\cite{ma2024agentboard}&None& Partial Obs. \& Temp. \& Stoch.  &None\\
\bottomrule
\end{tabular}
\end{threeparttable}
}
}
\end{table*}

While three distinct roles exist in a working system of agents, we argue that the agents should align with all the roles in a unified manner. To promote the orchestration of agents, humans, and the environment, the agents should work in the direction of eliminating the gap between agents and humans, agents and the environment, as well as adapting to the constraints imposed on the agents themselves. Based on this, we propose the principles of \textbf{U}nified \textbf{A}lignment for \textbf{A}gents (\textbf{UA}$^2$). Specifically, the agents should
\begin{enumerate}\setlength{\itemsep}{0pt}
    \item Align with \textit{human intentions}. The agents need to correctly recognize the intentions of the human users. While the goal is usually specified with a textual sentence, the ambiguity of language expression can affect the understanding and decision-making of agents.
    
    \item Align with \textit{environmental dynamics}. The agents need to interact with the environment to achieve the goal required by human users. To succeed in the achievement, the agents should raise their awareness of the operation laws of the environment. This is also advocated in~\cite{lecun2022path,hu2023language} that proposes to construct and incorporate a world model into an agent system.
    \item Align with \textit{self-constraints}. The underscored factor of current agent research and development comes from the constraints imposed on the agents themselves, including time and/or money budget limits. For foundation model agents, the underlying models (\textit{e.g.}, proprietary LLMs/LMMs) are costly for inference, which hurdles the performance of agents in realistic scenarios.
    
\end{enumerate}

In summary, the principles of \textbf{UA}$^2$ suggest that agents should achieve the unified alignment with \textit{human intentions}, \textit{environmental dynamics}, and \textit{self-constraints}. See Figure \ref{fig-2} as an illustration of the principles of \textbf{UA}$^2$.

\subsection{Challenges from the Principles of \textbf{UA}$^2$}
\label{a2challenges}

The principles of \textbf{UA}$^2$ have covered different sources of alignment for agents in a working system. In this section, we pose the special challenges raised from \textbf{UA}$^2$.

\textbf{Challenges in the alignment with \textit{human intentions}.} When the interaction between humans and the agent is a single-turn process, it is equivalent to LLM alignment~\cite{ouyang2022training} in the form of a prompt-response pair. However, in realistic settings, human intentions are often not perfectly covered in a single prompt, but rather reflected by preferences not directly visible from instructions (\textit{e.g.}, personal preferences and safety concerns). Challenges arise for the agents to infer authentic human intentions with multiple turns of interactions by either eliciting human preferences~\cite{li2023eliciting}, or learning to self-correct from environmental feedback~\cite{huang2023large}, or both.

\textbf{Challenges in the alignment with \textit{environmental dynamics}.} The interactive environments for agents in realistic scenarios can be highly complicated, which requires the agent to recognize the hidden state from the history of observations. Considering the dynamics function $\mathbf{s}_{n+1} \sim \pi(\mathbf{s}_n, \mathbf{a}_n)$ where $\mathbf{s}_n$ and $\mathbf{a}_n$ stand for the $n$-th step state and action, respectively, the complexity emerges from different parts:
\begin{itemize}\setlength{\itemsep}{0pt}
    \vspace{-5pt}
    \item Partial observability. This is reflected by the complexity of the function that transforms the historical observations $\{\mathbf{o}_{\le n}\}$ into the authentic state $\mathbf{s}_n$.
    \item Time-variant property. This is reflected by the temporal effect in the dynamics function, where the evolution of time $t$ leads to the variation of $\mathbf{s}_{n+1} \sim \pi(\mathbf{s}_n, \mathbf{a}_n, t)$.
    \looseness=-1 \item Stochasticity. The state transition of $\pi(\mathbf{s}_n, \mathbf{a}_n)$ can be interlaced with (nearly) independent random events.
    \vspace{-5pt}
\end{itemize}
In this way, constructing a precise world model for an agent system requires delicate techniques beyond ad-hoc exploration, coarse-grained memory, or ungrounded planning.

\textbf{Challenges in the alignment with \textit{self-constraints}.} The self-constraints of agents are the often-overlooked desiderata in the design of existing agent methodologies. Taking the budgetary limits (\textit{e.g.}, total time or number of tokens consumed in the foundation models) into account, the agent system should re-use the accumulated experiences during the lifelong learning process~\cite{majumder2023clin}, and balance the resources invested in the different learning modules. Furthermore, in scenarios where the self-constraints change with different episodes, additional challenges emerge for the agents to adapt to the constraints autonomously.

\section{Literature Review from the Lens of UA$^2$}

\subsection{Benchmarks}\label{sec-31}

In this section, we begin with a comprehensive review of current benchmarks in agent research, from the perspective of \textbf{UA}$^2$.
Representative benchmarks in both digital~\cite{toyama2021androidenv,yao2022webshop} and embodied~\cite{puig2018virtualhome, chevalierboisvert2019babyai} environments are summarized in Table~\ref{table:agents}.
By rendering realistic simulations~\cite{puig2023habitat,szot2021habitat} and carefully configured tasks~\cite{li2023behavior}, current benchmarks offer diverse environments for both language-based and embodied agents~\cite{xi2023rise} to operate and interact within~\cite{maes1995artificial}.
Instead of focusing on environmental authenticity~\cite{fu2022rfuniverse} or general task complexity, we assess the benchmarks prioritizing the alignment principles of \textbf{UA}$^2$.
In practice, we consider the following three aspects:

\begin{enumerate}\setlength{\itemsep}{0pt}
    \item \textit{Human intentions}: Whether the authentic goals need to be inferred during task execution, or the intentions of humans are precisely conveyed in the descriptions.
    \item \textit{Environmental dynamics}: Whether the state transitions of the environment are intrinsically endowed with partial observability, temporality, or stochasticity.
    \item \textit{Self-constraints}: Whether the status of budgetary resources is reflected, including time consumption, the maximum number of actions or reasoning steps, etc.
\end{enumerate}

\begin{figure*}[t]
    \centering
    \includegraphics[width=0.999\textwidth]{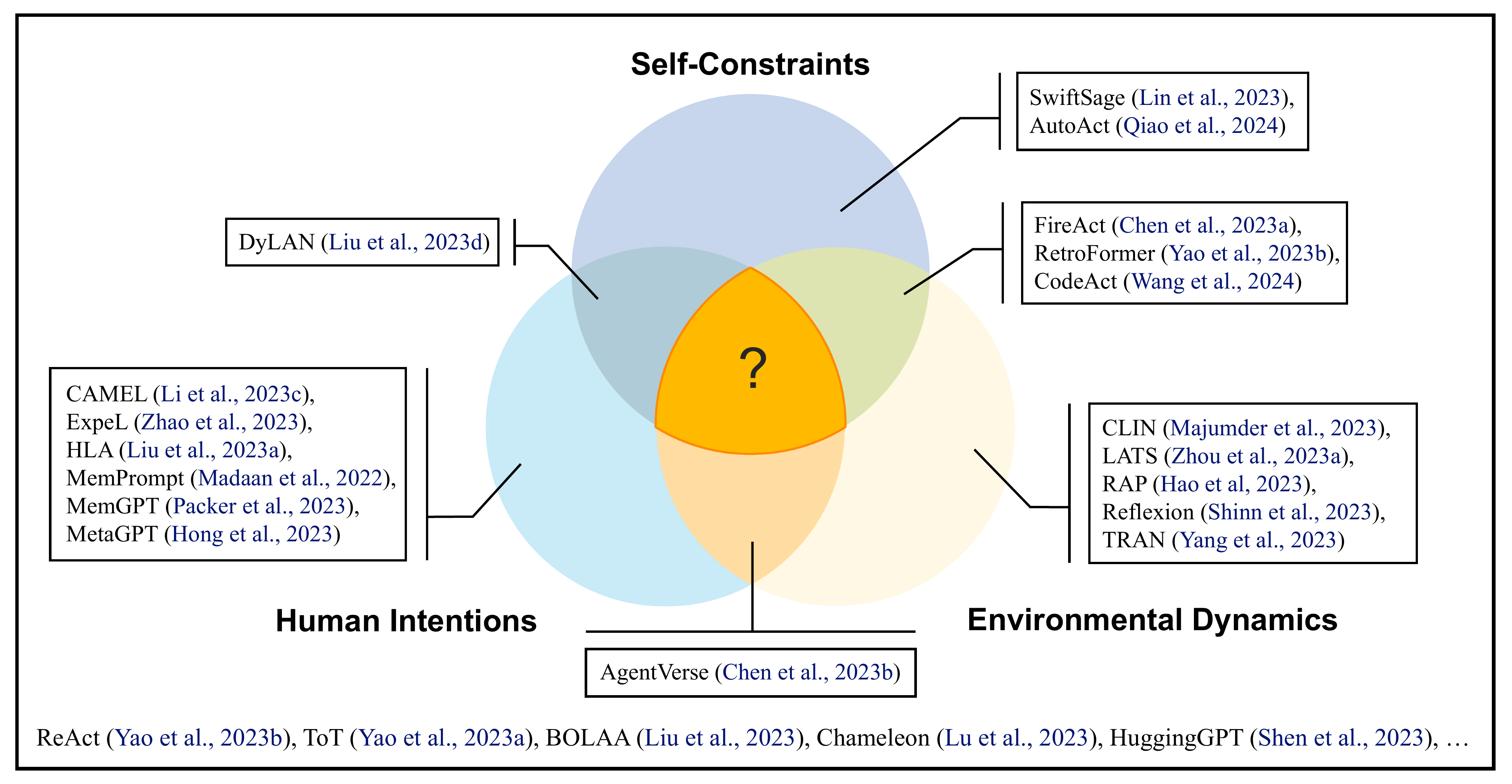}
    \vspace{-20pt}
    \caption{The dissection of alignment endeavors for different agent techniques. Generally speaking, the methods that actively coordinate with humans excel at aligning with \textit{human intentions}. The methods that are grounded with external feedback from the environment align well \textit{environmental dynamics}. The methods that adopt adaptive strategies or fine-tuning demonstrate better alignment with \textit{self-constraints}. While the advanced techniques mostly align with one role or two in the working system of agents, much room lies in the quest for \textbf{UA}$^2$.}
    \vspace{-5pt}
    
    \label{fig:review-methods}
\end{figure*}

In terms of human intentions, most benchmarks~\cite{qin2023toolllm, liu2023agentbench} provide explicit task instructions for more effective evaluation, rather than considering human intentions as hidden attributes for agents to discover.
By incorporating human interactions, several embodied simulators~\cite{puig2023habitat, xia2019gibson} facilitate tasks with vague goal descriptions~\cite{paul2022avlen, liu2023active}, necessitating agents to engage with humans to gather sufficient information for task completion.
In contrast, digital benchmarks hardly account for this aspect. The most relevant digital environment in this aspect is WebArena~\cite{zhou2023webarena}, which deliberately defines consistent human intentions across episodes. However, the intentions are also explicitly stated in the instructions, which bypasses the intention elicitation process of agents with humans. 

The benchmarks for agents are designed to mirror the complexities of the real-world dynamics~\cite{puig2023habitat}. Most benchmarks assume the environment is partially observable where agents are required to accomplish tasks through exploration and interaction~\cite{xia2020interactive}. Some benchmarks also include stochastic factors~\cite{wu2023smartplay,zhou2024hazard} or evolve with time~\cite{qin2023toolllm}.

Nevertheless, the synthesis of fine-grained realistic dynamics remains underdeveloped in benchmark design, resulting in the lack of evaluations of agent methodologies therein.

As for \textit{self-constraints}, embodied benchmarks~\cite{xia2020interactive,li2023behavior} use the number of actions as a metric to reflect the operational cost in real-world deployments, such as the path length in navigation tasks~\cite{anderson2018evaluation}.
In this context, AGENT~\cite{shu2021agent} further explicitly evaluates the trade-offs between cost and reward.
However, existing digital benchmarks mostly overlook cost and time constraints in the assessments (with \cite{wang2023mint} as an exception), which should be equally important.

In essence, existing agent benchmarks are still inadequate from the lens of \textbf{UA}$^2$. Furthermore, in general, the development of digital benchmarks lags behind that of embodied benchmarks. This underscores the need for the construction of more comprehensive and realistic environments, which could contribute to the advancement of agent techniques.

\subsection{Methods}\label{sec-32}

In this section, we review the representative agent methods from the perspective of \textbf{UA}$^2$. For each method, we analyze whether it actively seeks alignment with \textit{human intentions}, \textit{environmental dynamics}, or \textit{self-constraints}. 

To align with \textit{human intentions}, the agent methods should coordinate with humans through reasoning or experience summarization. HLA~\cite{liu2023llm} and MemPrompt~\cite{madaan2022memprompt} interact with humans for multiple rounds to solicit authentic human intentions. Multi-agent frameworks like CAMEL~\cite{li2023camel}, AgentVerse~\cite{chen2023agentverse}, and DyLAN~\cite{liu2023dynamic} leverage a group of agents for role-playing and inter-discussion to improve the understanding of human instructions. ExpeL~\cite{zhao2023expel} and MemGPT~\cite{packer2023memgpt} also align with human intentions through the analysis of human goals in an iterative manner.

To align with \textit{environmental dynamics}, the agents should ground themselves with external information from the environment. Reflexion~\cite{shinn2023reflexion}, LATS~\cite{zhou2023language}, and AgentVerse~\cite{chen2023agentverse} use external reward feedback as conditions to rectify their actions and improve the alignment with the environment. RetroFormer~\cite{yao2023retroformer}, TRAN~\cite{yang2023failures}, CLIN~\cite{majumder2023clin}, and FireAct~\cite{chen2023fireact} integrate the (abstracted) trajectories accumulated through the interaction with the environment into the prompts or data for fine-tuning. This results in an in-context or parametrized world model, which narrows the gap of alignment with the environment. RAP~\cite{hao2023reasoning} can also be categorized as aligning with the environment through simulation of the underlying foundation models.

To align with \textit{self-constraints}, the agents should adopt an adaptive strategy in the process of task execution and/or group construction. The representative works in this vein include SwiftSage~\cite{lin2023swiftsage}, Retroformer~\cite{yao2023retroformer}, and DyLAN~\cite{liu2023dynamic}. Finetuning a small-sized foundation model is also beneficial to the obedience of self-constraints~\cite{chen2023fireact,qiao2024autoact,wang2024executable}, which saves the cost of API calls of proprietary foundation models once the training is finished. 

In addition to the aforementioned frameworks, there are also basic techniques for agents, such as ReAct~\cite{yao2022react} and Tree-of-Thoughts~\cite{yao2023tree}, that serve as the foundational elements in most of the advanced agents. An overview of the analysis is illustrated in Figure~\ref{fig:review-methods}.

\looseness=-1 Despite the emergence of diverse agent techniques, plenty of room still exists for the unified alignment of agents with \textit{human intentions}, \textit{environmental dynamics}, and \textit{self-constraints}. Challenges lie in the construction of the agent framework~\cite{sumers2023cognitive}, which requires an elaborate design to strike a good balance of alignment with all three roles. Counterexamples in this sense are Reflexion~\cite{shinn2023reflexion} and LATS~\cite{zhou2023language}, which leverage multiple rounds of sampling to achieve better alignment with the environment, but the self-constraints are significantly violated at the same time due to the high cost. Moreover, the capability of the underlying foundation model dominates the potential of the sophisticated alignment endeavors of an agent. Therefore, it is essential to promote the synergy between the development of foundation models (such as alignment techniques) and the research of agents.

\section{Proof-of-Concept Studies}

In this section, we conduct proof-of-concept studies to validate the importance of \textbf{UA}$^2$ in the design of both benchmarks and methods for agents. \cref{sec-41} covers several realistic features we introduced into WebShop~\cite{yao2022webshop}, which are selected according to the principles of \textbf{UA}$^2$. In \cref{sec-42}, we introduce our agent method design following the principles of \textbf{UA}$^2$. \cref{sec-43} covers the experiments of several agent candidate baselines and our method in the retrofitted environment, and \cref{sec-44} reports the results as well as our discussions and findings.

\subsection{Environment Construction}\label{sec-41}

We conduct the case studies by first upgrading the WebShop environment. WebShop is a simulated online shopping environment with 1.18M real-world shopping items gathered from Amazon, and 12,087 textual shopping instructions collected from human annotators. While serving as a high-quality testbed for the instruction-following and planning abilities of foundation model agents, we further improve the complexity of WebShop by introducing the realistic factors around the three roles in the agent working system: \textit{human intentions}, \textit{environmental dynamics}, and \textit{self-constraints}.

\textbf{Human intentions.} In reality, different human users own unique, potentially invisible preferences about the properties and categories of shopping items. Given this, we configure 10 different users for testing, each possessing a basic preference (in text) that corresponds with a certain hidden attribute of items. We equip each user with a group of 50 consecutive artificially constructed instructions with user profiles, ambiguous descriptions, and preferences to be inferred by tracking the purchase history. The rules of reward computation for each instruction follow those of the original WebShop (see Appendix~\ref{appendix-1-1} for details).

\textbf{Environmental dynamics.} To narrow the gap with realistic online shopping scenarios, we implement fine-grained personalized reranking algorithms on top of the original search engine in WebShop. The algorithms include collaborative filtering~\cite{collaborative} and a Determinantal Point Process (DPP) based method~\cite{NEURIPS2018_dbbf603f}. The environment then constantly evolves with user actions and reflects the complexity of realistic environmental dynamics. The details of the implementation are listed in Appendix~\ref{appendix-1-2}.

\textbf{Self-constraints.} To measure the expenses of the agents themselves during the operating process, we implement the runtime environment to count for the temporal and monetary expenditures for the agent working system. The monetary cost consists of the API calls of the proprietary foundation models, and the time consumption indicates the normalized endurance of interaction between the agents and the interactive environment (detailed in Appendix~\ref{appendix-1-3}).

\subsection{Agent Design with the Principles of \textbf{UA}$^2$}\label{sec-42}

\begin{figure}[t]
    \centering
    \includegraphics[width=\linewidth]{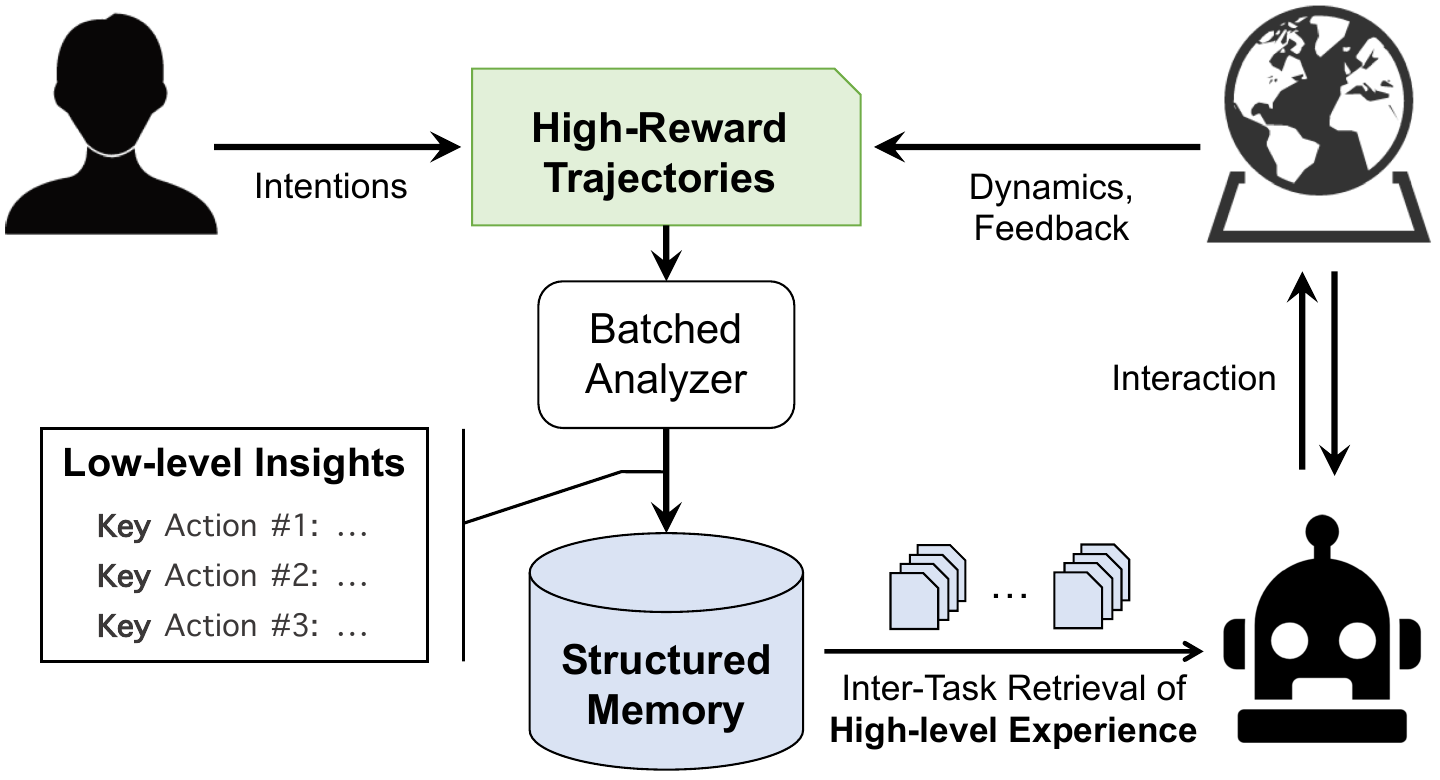}
    \caption{Overview of our agent design that follows the principles of \textbf{UA}$^2$. By continually analyzing and retrieving structured memory from similar tasks of the same user, the agent manages to extrapolate past experience and generalize across different tasks.}
    \label{fig:method-a2agent}
\end{figure}

\begin{table*}[t]
    \caption{The performance of averaged reward, success rate (SR) (\%),  alignment gap (\%) with human intentions ($\mathbf{G}_{\mathrm{HI}}$) and environment dynamics ($\mathbf{G}_{\mathrm{ED}}$), time (s) and money (\$) cost of all methods tested in our retrofitted WebShop environment. The best result for each metric is in \textbf{bold}. The better performance under each metric is indicated by the darker \textcolor{mycolor4}{green} shades. *LATS is tested on 1/10 subset of the entire task instructions due to the significant cost.}
    \label{tab:results}
    \centering
    \begin{tabular}{l|cc|cccc}
    \toprule
    Method & Reward $\uparrow$ & ~~SR (\%) $\uparrow$ & $\mathbf{G}_{\mathrm{HI}}$ (\%) $\downarrow$ & $\mathbf{G}_{\mathrm{ED}}$ (\%) $\downarrow$ & ~~~~Time (s) $\downarrow$ & Money (\$) $\downarrow$ \\
    \midrule
    ReAct      & \cellcolor{mycolor2}50.3 & \cellcolor{mycolor2}~~8.0 & \cellcolor{mycolor2}11.7 & \cellcolor{mycolor3}14.9 & \cellcolor{mycolor4}~~~~\textbf{1.716} & \cellcolor{mycolor4}\textbf{0.013} \\
    ReAct-SC   & \cellcolor{mycolor2}49.9 & \cellcolor{mycolor2}~~7.4 & \cellcolor{mycolor1}14.4 & \cellcolor{mycolor3}14.6 & \cellcolor{mycolor3}~~~~1.720 & \cellcolor{mycolor2}0.039 \\
    Reflexion  & \cellcolor{mycolor1}44.4 & \cellcolor{mycolor4}\textbf{13.8} & \cellcolor{mycolor1}22.5 & \cellcolor{mycolor1}25.7 & \cellcolor{mycolor2}~~~~5.539 & \cellcolor{mycolor2}0.045 \\
    LATS*       & \cellcolor{mycolor4}\textbf{52.4} & \cellcolor{mycolor3}10.0 & \cellcolor{mycolor1}18.5 & \cellcolor{mycolor4}\textbf{14.3} & \cellcolor{mycolor1}125.935 & \cellcolor{mycolor1}5.508 \\
    \midrule
    Ours       & \cellcolor{mycolor3}51.9 & \cellcolor{mycolor3}~~9.6 & \cellcolor{mycolor4}~~\textbf{6.7} & \cellcolor{mycolor3}14.8 & \cellcolor{mycolor3}~~~~1.779 & \cellcolor{mycolor3}0.014 \\
    \bottomrule
    \end{tabular}
\end{table*}

Following the principles of \textbf{UA}$^2$, we initiate our agent by introducing the structured memory module on top of ReAct~\citep{yao2022react}. As shown in Figure~\ref{fig:method-a2agent}, the introduced module is formed by two key components: \textit{low-level} action insights and \textit{high-level} intra-task experience.

\looseness=-1 \textit{Low-level} action insights are a list of key actions solicited from different runs in the environment under the same task instruction. The key actions are extracted from the high-reward trajectories with an analyzer, with which the contributions of actions are computed in the task-solving process. The analyzer adopts a batched inference \citep{cheng-etal-2023-batch} to tag all actions at the same time. The structured memory is then maintained with the accumulation of key actions paired with their corresponding human instructions.

\textit{High-level} intra-task experience is formed by the retrieval of the \textit{low-level} action insights accumulated in the structured memory. According to the similarity of the current human instruction with the previous ones stored in the memory, the key actions are gathered to form an initial plan for the current task. The re-use of high-level experience throughout the stream of tasks promotes efficient intra-task generalization.

We design the agent to differ from previous works, which rely on LLM summarization of unstructured insights~\citep{majumder2023clin,zhao2023expel} or multiple-round LLM reflections within a single task~\citep{shinn2023reflexion}. Our method aligns with \textit{human intentions}, \textit{environmental dynamics}, as well as \textit{self-constraints}: (i) The maintenance of the structured memory contributes to the lifelong profiling of a human user. (ii) The storage and retrieval of key actions analyzed from different trajectories improves the awareness of the agent to the environment. (iii) The reuse of structured records saves the agents from planning from scratch for each task, which aligns with \textit{self-constraints} by cost minimization. \cref{app:exp} covers the formal descriptions and implementation details of our method.

\subsection{Experiments}\label{sec-43}

\paragraph{Baselines.}

We compare the performance of our method with several widely-used agent techniques on the retrofitted WebShop in Section~\ref{sec-41}, including (1) ReAct~\cite{yao2022react}, which harmonizes internal reasoning and external actions, (2) ReAct-SC (ReAct with Self-Consistency), which equips ReAct with sampling and marginalization~\cite{wang2022self}, (3) Reflexion~\cite{shinn2023reflexion}, which conducts self-correction by reflecting on past actions and observations, and (4) LATS~\cite{zhou2023language}, which leverages a combination of techniques including ReAct, self-reflection, and Monte Carlo Tree Search (MCTS). Note that we leave the implementation of techniques categorized as aligning with human intentions in Section~\ref{sec-32} as future work, since great effort should be taken by involving humans in the interaction loop and adapting to our settings.

\paragraph{Evaluation Metrics.}

\looseness=-1 Following the settings of \citet{yao2022webshop}, we measure the performance of task completion with the average reward and success rate incurred per task. To quantitatively investigate the alignment of different methods under the principles of \textbf{UA}$^2$, we introduce three extra metrics. We report the averaged monetary and time cost to reflect the alignment of each method with \textit{self-constraints}. For \textit{human intentions} and \textit{environmental dynamics}, we build ablated versions of the retrofitted WebShop that exclude the introduced feature, respectively. We then test each agent technique on the pair of fully-retrofitted / ablated environments, and finally investigate the difference between the pair of the evaluated rewards. More specifically:

To evaluate the alignment with \textit{human intentions}, we construct an ablated version of the environment in \cref{sec-41}, where the hidden attributes corresponding with user profiles or preferences are excluded from the reward computation. In this ablated environment, the performance of each method should be better than that in the fully-retrofitted environment. We define the alignment gap with human intentions $\mathbf{G}_{\mathrm{HI}}$ as the relative difference between the two performances:
\begin{equation}
    \mathbf{G}_{\mathrm{HI}} = (R_{\mathrm{full}} - R_{\mathrm{HI}}) / R_{\mathrm{full}} \times 100\%,
\end{equation}
where $R_{\mathrm{full}}$ and $R_{\mathrm{HI}}$ stand for the reward of an agent in the fully-retrofitted environment and the environment excluding the computation of human intentions, respectively.

\looseness=-1 Similarly, to evaluate the alignment with \textit{environmental dynamics}, we build an ablated environment without the implementation of the personalized reranking algorithms, and define the alignment gap with environmental dynamics $\mathbf{G}_{\mathrm{ED}}$:
\begin{equation}
    \mathbf{G}_{\mathrm{ED}} = (R_{\mathrm{full}} - R_{\mathrm{ED}}) / R_{\mathrm{full}} \times 100\%,
\end{equation}
with $R_{\mathrm{ED}}$ as the reward of an agent in the ablated environment that excludes the personalized reranking algorithms.

\subsection{Results and Discussions}\label{sec-44}

The performances of different methods in all the metrics are shown in Table \ref{tab:results}. According to the results, Our framework achieves the top unified performance among all the methods, with the best balance between task completion performance and measures of different alignment sources.

LATS achieves the highest average reward, and Reflexion obtains the top success rate. This is because they both employ trial-and-error approaches with multiple rounds of interactions.
However, the money and time costs of the two methods are significantly higher than other methods, suggesting their weaknesses in aligning with the \textit{self-constraints} of agents.  
To be specific, Reflexion incurs a cost over 5$\times$ in time and 3$\times$ in money compared to other methods, while LATS, in contrast with other methods, entails a cost exceeding 100$\times$ in time and nearly 200$\times$ in money.

ReAct-SC achieves a comparable average reward and success rate (SR) with ReAct. This might be attributed to the complexity of our retrofitted environment, where even more runs of sampling are required in ReAct-SC to vote for better actions. In addition, The incorporation of self-consistency in ReAct-SC requires more calls of the API of the proprietary foundation model, resulting in approximately three times the cost of money compared to ReAct. The time cost of ReAct and ReAct-SC is nearly identical. This is because we only document the endurance within the interactive environment (the time of API requests is neglected), and at the same time, ReAct might exhibit similar planning abilities as ReAct-SC. Finally, Our framework achieves the top performance in averaged rewards and success rates, which underscores the significance of the principles of \textbf{UA}$^2$.

As for the alignment gap, the results of $\mathbf{G}_{\mathrm{HI}}$ and $\mathbf{G}_{\mathrm{ED}}$ in Table \ref{tab:results} indicate that almost all baselines possess the gap above 10\% in terms of aligning with humans or the complex environment. Notably, our method demonstrates a significantly lower $\mathbf{G}_{\mathrm{HI}}$ than other methods, which might benefit from its capacity to adapt to diverse human intentions. In contrast, LATS demonstrates a relatively low $\mathbf{G}_{\mathrm{ED}}$ of 14.3\%. This is because of the accumulation of trials from the exhaustive sampling in the environment, which also limits its practical applicability. For comparison, neither $\mathbf{G}_{\mathrm{HI}}$ nor $\mathbf{G}_{\mathrm{ED}}$ of Reflexion is satisfactory, which might indicate that the mechanism of the self-reflection is inferior to other techniques in this setting.
These results highlight the need for agent techniques following the principles of \textbf{UA}$^2$.

\section{Actionable Insights}

Envisioning the future of autonomous agents powered by foundation models in real-world applications, in this section, we provide insights on the next steps of research from \textbf{UA}$^{2}$.

\textbf{Synergizing agents with alignment research.} Alignment research aims to steer a model to follow instructions faithfully. To achieve unified alignment in an agent system, techniques in the field of alignment research can be helpful to the foundation model agents in following the principles of \textbf{UA}$^2$. For instance, humans can leverage ideas like Constitutional AI~\cite{bai2022constitutional} to integrate the principles of unified alignment into the objectives of the agents.

\textbf{Constructing realistic agent benchmarks.} While appreciating the existing efforts on the benchmark construction for agents, we advocate for more realistic simulation and sandbox design reflecting the intricate scenarios with nuanced logistics and details. As shown in our proof-of-concept studies, the principles of \textbf{UA}$^2$ are also helpful in the design of the benchmark. Taking \textbf{UA}$^2$ into account, the gap between agents and realistic human demands and interactive environment can be revealed more faithfully, laying the foundation for the next breakthrough of agent techniques.

\textbf{Developing holistic evaluations for agents.} Existing research on agents mainly focuses on the final success of task completion. In our work, we propose the principles of unified alignment for agents, suggesting the proficiency of agents can be reflected by the quality of alignment with \textit{human intentions}, \textit{environmental dynamics}, and \textit{self-constraints}. Given this, the dissection of the performance of agents is necessary for the development of agent techniques, since an analysis of alignment gaps with different roles indicates the direction of improvement for agents. This suggests the significance of holistic evaluations for agents.

\textbf{Toward self-evolving agents through continual alignment.} While the sources of alignment have been categorized by \textbf{UA}$^2$, it requires the elaborated design of agent methods that carefully balance the different alignment sources in a unified manner. Envisioning agents with next-level autonomy, we expect the agents to self-evolve through lifelong interaction with humans and the environment with continual alignment. In this vein, agents improve themselves with better use and efficiency, leading to general problem-solving abilities in complex, real-world scenarios.

\vspace{-2pt}
\section{Conclusion}

In this work, we propose the principles of unified alignment for agents with \textit{human intentions}, \textit{environmental dynamics}, and \textit{self-constraints}. We start by recognizing the three components in a working system of agents: agents, humans, and environment, then state the necessity of agents to align with the three roles in a unified manner and propose the principles of \textbf{UA}$^2$. We demonstrate the significance of \textbf{UA}$^2$ by literature review and proof-of-concept studies. Eventually, we shed light on the impact of \textbf{UA}$^2$ on the future of agent research with enhanced general problem-solving abilities.

\section*{Impact Statement}

The prosperity of autonomous agents with foundation models has posed exciting avenues for future research toward the automatic execution of daily tasks for humans. In our work, we advocate for the unified alignment of agents (\textbf{UA}$^2$) with humans, the environment, and the agents themselves simultaneously. To align with humans means to improve the understanding of \textit{human intentions}, and especially safety concerns, to provide better assistance. By doing so, the agents also need to align with the environment to enhance the awareness of \textit{environmental dynamics}, so that the agents can be cautious about whether the next actions could be malicious or destructive. The agents should also align with themselves in terms of \textit{self-constraints}, adhering to the running cost of money, time, battery, etc. In our work, we conduct proof-of-concept studies by introducing realistic features, such as human profiles, personalized reranking algorithms, and runtime cost counters into the original WebShop. While the results have proved the essence of \textbf{UA}$^2$, we plan to experiment with extra alignment factors in the future, including safety concerns from human intentions, random events from the environment, and others.

Our work covers the principles for agents to follow, and we expect the future of agents with narrowed alignment gaps in a unified manner. We also expect the construction of more realistic sandboxes or simulators as the testbeds for agents, where both the capability and safety of agents can be better studied and improved under realistic settings. Eventually, our principles of unified alignment for agents lay the foundation for the next-level autonomous agents more intelligent and more responsible.

\section*{Contributions}

\subsection*{1. Proof-of-Concept: Environment Construction}

\textbf{Zonghan Yang}: \textit{Led and initiated environment construction according to \textbf{UA}$^2$}; \textit{Designed the test tasks with user profiles and instruction groups};

\textbf{Fangzhou Xiong}: \textit{Implemented the two reranking algorithms on top of the original WebShop};

\textbf{Yile Wang}: \textit{Led and implemented ChatGPT role-playing to gather simulated preference data for collaborative filtering};

\textbf{Kaiming Liu}, \textbf{Zonghan Yang} and \textbf{Zeyuan Yang}: \textit{Implemented the runtime budgetary counter};

\textbf{Xinrui Chen}: \textit{Implemented the user login mechanism to isolate different test agents};

\textbf{Zhenhe Zhang}: \textit{Estimated the response delays of the website for the self cost counter};

\textbf{Xinrui Chen} and \textbf{Zhenhe Zhang}: \textit{Gathered and processed the simulated preference data};

\textbf{All the aforementioned}: \textit{Contributed to the validity checks of the task construction};

\textbf{Chi Chen}, \textbf{Fuwen Luo}, \textbf{Ziyue Wang}, \textbf{Siyu Wang}, and \textbf{Xiaolong Wang}: \textit{Contributed to the human evaluation of the ChatGPT-simulated preference data};

\subsection*{2. Proof-of-Concept: Agent Design and Experiments}

\textbf{An Liu}: \textit{Co-led the experimental analysis; Design and contributor of our agent framework according to the principles of \textbf{UA}$^2$; Main implementation of Reflexion and LATS};

\textbf{Zijun Liu}: \textit{Co-led the experimental analysis; Main design and contributor of our agent framework};

\looseness=-1 \textbf{Kaiming Liu}: \textit{Thorough experiments and ablation studies}; \textit{Contributor of our agent framework}; \textit{Main implementation of the ReAct family and the CoT family};

\textbf{Qingyuan Hu}: \textit{Implementation of the CoT family};

\textbf{Zhicheng Guo}: \textit{Implementation of LATS};

\textbf{Zonghan Yang}: \textit{Initial implementation of ReAct};

\subsection*{3. Writing (Main Paper)}

\textbf{Draft of Section 3}: Section 3.1: Zeyuan Yang, Yile Wang, Zhicheng Guo, and Fuwen Luo; Section 3.2: Zijun Liu, Kaiming Liu, and Zeyuan Yang;

\textbf{Draft of Section 4}: Section 4.1: Zonghan Yang, Yile Wang, Fangzhou Xiong, and Kaiming Liu; Section 4.2: Zijun Liu, and Kaiming Liu; Sections 4.3 and 4.4: An Liu, Kaiming Liu, Zhicheng Guo, and Qingyuan Hu.

\textbf{Draft of Other Parts and Finalization}: Zonghan Yang.

\subsection*{4. Additional Contributions}

\textbf{Project Supervision}: Peng Li and Yang Liu;

\textbf{Feedback for Early Versions of Paper Drafts}: Xiaoyue Mi, Xuanyu Lei, Ziyue Wang, Peng Li, and Yang Liu;

\textbf{Early Discussions}: Chi Chen and Sijie Cheng;

\textbf{Figure helps}: Ziwei Chi.

\bibliography{agentforce}
\bibliographystyle{icml2024}

\newpage
\appendix
\onecolumn

\section{Environment Construction in Section \ref{sec-41}} \label{appendix-1}

In this section, We introduce the realistic features we introduce into the original WebShop in detail. Note that as we aim to conduct \textit{proof-of-concept} studies, the features are implemented for the purpose of \textit{reflecting the three lines of alignment only}. We also anticipate realistic configurations and more nuanced logistics in a dedicated benchmark in the future.

\subsection{Task Design in the retrofitted WebShop} \label{appendix-1-1}

Different from the precise human instructions in the original WebShop environment, we design tasks to reflect the necessity of agents to align with \textit{human intentions}. In reality, different human users own unique preferences about the properties and categories of shopping items. Such preferences form the profile of a user, which dominates their authentic intentions in the stream of shopping instructions. As it is not always easy for human users to explicitly write down the precise instructions for all their shopping intentions, the agent should assist humans in tracking and inferring human intentions in a lifelong manner. 

Given this, we configure 10 different users, each possessing a basic preference (described in text) that corresponds with a certain hidden attribute of items. For example, for the hidden attribute \texttt{cruelty-free}, we design the corresponding basic human profile sentence to be \texttt{I cannot care enough for the cute creatures in this world.} We follow the reward computation rules in the original WebShop, therefore the match of hidden attributes is essential to the final reward. We equip each user with a group of 50 consecutive artificially constructed instructions. In all the instructions of the group, the aforementioned profile sentence always appears. 

The specific instruction for the purchase of this round falls into three cases: 
\begin{itemize}
    \vspace{-5pt}
    \item The basic preference of the user should be considered. An example in this category reads: \texttt{i am interested in a 60 count of toner that is suitable for sensitive skin, and price lower than 50.00 dollars.} For this instruction, the expected item to be found contains two hidden attributes: \texttt{sensitive skin} according to the task instruction, and \texttt{cruelty-free} according to the user profile. 
    \item The basic preference of the user does not need to be considered. An example in this category reads: \texttt{i am looking for a nightstand that is easy to install.} Our motivation for this case is that the users with whatever preferences always have the need to buy some specific items, which can be irrelevant to their profiles. In this example, the set of the underlying hidden attributes contains just the task-related \texttt{easy install}.
    \item The basic preference of the user should be considered, \textit{and} an extra preference should be recognized according to the recent instruction histories. An example in this category reads: \texttt{Based on my purchase preference from history, help me to buy eco friendly face towels.} And the corresponding instruction history for this example is listed in the following Table \ref{table:ins-his}. It should be inferred according to the history that the hidden attribute (as the invisible intention) that frequently appears in recent instructions are \texttt{sensitive skin}. We design the order of the instruction group so that such to-be-inferred attributes appear at least five times more than other attribute candidates in a sliding window. As a result, in this example, the set of the hidden attributes consists of the \texttt{cruelty free} according to the user profile, the \texttt{eco friendly} according to the description in the textual instruction, and also the \texttt{sensitive skin} that is to be tracked and inferred from history.
    \begin{table}[h!]
    \centering
    \small
    \vspace{-10pt}
    \caption{The instruction history for the example.}
    \begin{tabular}{l}
    \toprule
    \textbf{The following instruction history is listed in reverse chronological order}: \\
    ~~~~i am interested in a 60 count of toner that is suitable for sensitive skin. \\
    ~~~~i am looking for a sulfate \& paraben free shampoo that is also suitable for my sensitive skin. \\
    ~~~~i want some hand cream for dry and sensitive hands in a grapefruit scent. \\
    ~~~~i need men's non toxic deororizing body spray for sensitive skin. get the 2pack 3.4 ounce size. \\
    ~~~~get me a body scrub to remove dead skin. pick a 3.4 fl oz pack that is meant for sensitive skin. \\
    ~~~~i need low rise boot cut pants in grey color. \\
    ~~~~buy me some paraben free coconut oil lip balm for my sensitive skin. \\
    ~~~~i'm looking for some juicy watermelon lip gloss that is paraben and oil free and suitable for sensitive skin. \\
    ~~~~i need soaps for dry and sensitive skin that are made with argan oil. \\
    ~~~~i'm looking to buy a body wash that has tea tree oil as an ingredient that would work well for sensitive skin. \\
    \bottomrule
    \end{tabular}
    \vspace{-5pt}
    \label{table:ins-his}
\end{table}
\end{itemize}

Most of the task instructions in the former two groups are selected from the crowdsourced instructions, whose corresponding ground-truth items are labeled with the hidden attributes of both the basic user preference and the task-related instruction. We artificially rewrite the instructions in the third category by introducing indicating words like \texttt{Based on my purchase from the history}. In the 10 groups of 50 consecutive instructions, the statistics of the three categories is 298/97/105 for the first/second/third category, approximately 3:1:1. The task completion performances of all agent techniques should be tested on the 10 groups of 50 instructions each, with the overall averaged reward and success rate reported. In the ablated version of the retrofitted environment for the calculation of $R_{\mathrm{HI}}$ and $\mathbf{G}_{\mathrm{HI}}$ in Section \ref{sec-43}, all the hidden attributes about the basic user profiles (\textit{e.g.}, the \texttt{cruelty free}) and the preferences to be inferred (\textit{e.g.}, the \texttt{sensitive skin}) are excluded from the reward computation.

\subsection{Personalized Reranking Algorithms} \label{appendix-1-2}

\subsubsection{Overview}

To narrow the gap with realistic online shopping scenarios, we implement personalized reranking algorithms in WebShop. With the reranking algorithms, the search results of the shopping item list are re-ordered according to the click histories of the users. Specifically, on top of the Pyserini~\cite{Lin_etal_SIGIR2021_Pyserini} search engine used in the original Webshop, we integrate a Collaborative Filtering (CF)~\cite{collaborative} algorithm and a Determinantal Point Process (DPP) based algorithm~\cite{NEURIPS2018_dbbf603f} for fine-grained personalized re-ranking. The DPP-based algorithm provides the reranking weights based on the historical actions of the agent itself, while the CF-based algorithm provides the reranking weights based on the similarity with other users. The two sets of weights by the two algorithms are eventually linearly averaged with the coefficient of $0.2$ for DPP-based weights and $0.8$ for CF-based weights. Driven by the reranking algorithms, the environment is constantly evolving with user actions, which could better reflect the complexity of realistic environmental dynamics. In the ablated version of the retrofitted environment for the calculation of $R_{\mathrm{ED}}$ and $\mathbf{G}_{\mathrm{ED}}$ in Section \ref{sec-43}, the two algorithms are disabled.

The algorithms of collaborative filtering and DPP-based reranking are briefed in Algorithms \ref{alg:cf} and \ref{alg:dpp}, respectively.

\begin{algorithm}
    \renewcommand{\algorithmicrequire}{\textbf{Input:}}
    \renewcommand{\algorithmicensure}{\textbf{Output:}}
    \caption{User-Based Collaborative Filtering}
    \label{alg:cf}
    \begin{algorithmic}[1]
        \REQUIRE Prime user-item rating matrix $\mathbf{R}$, User Click Through Rate $\mathbf{U}$, Top-n items $\mathbf{I}$
        \ENSURE  CF reranking score of top-n items $\mathbf{Y}$
        \FOR{each prime user i}
            \FOR{$j \in \mathbf{I}$}
                \STATE $\mathbf{M}_{i, j} = \mathbf{R}_{i, j}$
            \ENDFOR
        \ENDFOR
            
        \FOR{each prime user i}
          \STATE \textcolor{blue}{/* Calculate the intersection of items contained in the current agent and prime users */ } \\
          \STATE $P = \mathbf{U} \cap \mathbf{R}_i$  \\ 
          \STATE $\mathbf{S}_i = \frac{\sum_{p \in P} (\mathbf{R}_{i, p} - \overline{\mathbf{R}}_i)(\mathbf{U}_{p} - \overline{\mathbf{U}})}{\sqrt{\sum_{p \in P} (\mathbf{R}_{i, p} - \overline{\mathbf{R}}_i)^2}\sqrt{\sum_{p \in P} (\mathbf{U}_{p} - \overline{\mathbf{U}})^2}}$
        \ENDFOR
        \STATE $\mathbf{Y} =  \mathbf{S} \mathbf{M}/\sum{\mathbf{S}_i}$
    \end{algorithmic}
\end{algorithm}

\begin{algorithm}
    \renewcommand{\algorithmicrequire}{\textbf{Input:}}
    \renewcommand{\algorithmicensure}{\textbf{Output:}}
    \caption{Deterministic Point Process Based Reranking~\cite{NEURIPS2018_dbbf603f}}
    \label{alg:dpp}
    \begin{algorithmic}[1]
        \REQUIRE Item score vector $\mathbf{I}$, Item similarity matrix $\mathbf{S}$, Top K\\
        \ENSURE  DPP-based reranking score of top-n items $Y_g$
        \STATE $\mathbf{c}_i = [], d_i^2 = \mathbf{L}_{ii}, j = \arg\max_{i \in Z} \log(d_i^2), Y_g = \{j\}$ \\ 
        \STATE $\mathbf{L} = \mathrm{diag}(\mathbf{I})~\mathbf{S}~\mathrm{diag}(\mathbf{I})$\\ 
        \WHILE{$\lvert Y_g \rvert < K$ }
            \FOR{$i \in Z \setminus Y_g$}
                \STATE $e_i = (\mathbf{L}_{ji} - \langle\mathbf{c}_j, \mathbf{c}_i\rangle) / d_j$ \\
                \STATE $\mathbf{c}_i = \mathbf{c}_i\|e_i$ \\
                \STATE $d_i^2 = d_i^2 - e_i^2$
            \ENDFOR
            \STATE $j = \arg\max_{i \in Z \setminus Y_g} \log(d_i^2), Y_g = Y_g \cup \{j\}$
        \ENDWHILE
    \end{algorithmic}
\end{algorithm}

In the implementation of the CF-based algorithm, we first employed ChatGPT (\texttt{gpt-3.5-turbo-1106}) as an assistant to simulate 30 different users and gather their preference data for collaborative filtering (detailed in Appendix \ref{app:user-behavior-simulation}). During the shopping process, we record the click-through rates (CTR) of the agent on every item. We then re-rank the item list of the search results according to the agent's CTR and its Pearson correlation between the simulated user preferences.

\subsubsection{User Behavior Simulation}\label{app:user-behavior-simulation}

We employed ChatGPT to design 30 different roles, simulating the process of shopping and gathering the ranking results for 50 products for each role, the prompts are shown in Table~\ref{table:user_simulation}. The collected data will be used to simulate user behavior for simulating the dynamic environment similar to recommendation systems with changeable displayed items for different users and behaviors. The information of 30 roles is shown in Table~\ref{table:roles_part1} and Table~\ref{table:roles_part2}. Note that these roles are completely generated by ChatGPT, including their genders and other information. 
In this work, we construct the 30 roles to obtain the set of preference weights for only the purpose of introducing the reranking mechanisms into the environment. We plan to refine the construction of the profiles with a broader coverage of demographic groups in the future.
\begin{table}[h!]
    \centering
    \small
    \caption{The prompt for user behavior simulation using ChatGPT. We define roles (colored in \textcolor{teal}{green}) and the query (colored in \textcolor{blue}{blue}), asking for reranking items (colored in \textcolor{orange}{orange}) given by the original Webshop. The generated results (colored in \textcolor{magenta}{red}) can serve as a simulation of user click behavior.}
    \begin{tabular}{l}
    \toprule
    \textbf{User Behavior Simulation} \\
    \midrule
    \textcolor{gray}{/* Prompt */} \\
    Your name is \textcolor{teal}{$[$\texttt{NAME}$]$} and here is your profile:  \\
    Gender: \textcolor{teal}{$[$\texttt{Gender}$]$}\\
    Age: \textcolor{teal}{$[$\texttt{Age}$]$}\\
    Occupation: \textcolor{teal}{$[$\texttt{Occupation}$]$}\\
    Shopping Habits: \textcolor{teal}{$[$\texttt{Shopping Habits}$]$}\\
    \\
    You are searching for \textcolor{blue}{$[$\texttt{Query}$]$} on a shopping website and obtain 50 results:\\
    \\
    Id: \textcolor{orange}{$[$\texttt{Id}$_1$$]$}; Description: \textcolor{orange}{$[$\texttt{Desc}$_1$$]$}; Price: \textcolor{orange}{$[$\texttt{Price}$_1$$]$}\\
    Id: \textcolor{orange}{$[$\texttt{Id}$_2$$]$}; Description: \textcolor{orange}{$[$\texttt{Desc}$_2$$]$}; Price: \textcolor{orange}{$[$\texttt{Price}$_2$$]$}\\
    ...\\
    Id: \textcolor{orange}{$[$\texttt{Id}$_{50}$$]$}; Description: \textcolor{orange}{$[$\texttt{Desc}$_{50}$$]$}; Price: \textcolor{orange}{$[$\texttt{Price}$_{50}$$]$}\\
    \\
    Please sort all 50 products according to your preferences using the format of ``Id-Ranking''.\\
    \\
    \textcolor{gray}{/* Response */} \\
    \textcolor{magenta}{$[$\texttt{Id}$_1$$]$-$[$\texttt{Rank}$_1$$]$}; \textcolor{magenta}{$[$\texttt{Id}$_2$$]$-$[$\texttt{Rank}$_2$$]$}; ...; \textcolor{magenta}{$[$\texttt{Id}$_{50}$$]$-$[$\texttt{Rank}$_{50}$$]$};\\
    \bottomrule
    \end{tabular}
    \label{table:user_simulation}
\end{table}

\begin{table*}[t]
\centering
\small
\caption{The generated 30 different simulated users by using ChatGPT (part 1).}
\begin{tabular}{lp{2.5cm}p{2.3cm}p{3cm}p{2.8cm}p{2.2cm}}
\toprule
\textbf{Profile}&\textbf{Role \#1}&\textbf{Role \#2}&\textbf{Role \#3}&\textbf{Role \#4}&\textbf{Role \#5}\\
\midrule
Name&Sarah&Juan&Lisa&Michael&Emma\\
Gender&Female&Male&Female&Male&Female\\
Age&32&21&40&45&55\\
Occupation&Software Engineer&College Student&Stay-at-home mom&Construction Manager&Retired Teacher\\
Habits&Sarah loves online shopping for the latest gadgets and tech accessories. She researches extensively, reads reviews, and compares prices before making a purchase. She's always on the lookout for the newest tech trends.&Juan is passionate about fashion and enjoys shopping for trendy clothing and accessories. He follows fashion influencers on social media, visits local boutiques, and regularly updates his wardrobe to stay stylish on campus.&Lisa prioritizes her family's health and wellness. She shops for organic groceries, supplements, and natural skincare products. She also invests in fitness equipment and enjoys trying new workout routines.&Michael is passionate about home improvement projects. He frequently visits hardware stores, researches tools and materials, and enjoys renovating and enhancing his living space. He seeks quality products for long-term durability.&Emma is an avid reader and loves hosting book club meetings. She enjoys browsing bookstores, collecting literary classics, and exploring various genres. She values recommendations from fellow book lovers.\\
\midrule
\textbf{Profile}&\textbf{Role \#6}&\textbf{Role \#7}&\textbf{Role \#8}&\textbf{Role \#9}&\textbf{Role \#10}\\
\midrule
Name&Alex&Ryan&Maya&Daniel&Olivia\\
Gender&Non-binary&Male&Female&Male&Female\\
Age&27&35&28&50&42\\
Occupation&Etsy Shop Owner&Chef&Environmental Scientist&Pet Store Owner&Financial Analyst\\
Habits&Alex loves creating unique handmade items and runs an online store. They actively seek out specialty craft supplies, materials, and tools to produce high-quality products. They also enjoy attending craft fairs and networking with other artisans.&Ryan is passionate about cooking and constantly seeks out new ingredients and culinary tools. He enjoys shopping at local markets, specialty food stores, and online platforms for gourmet products. He values quality and freshness.&Maya loves hiking, camping, and exploring nature. She invests in high-quality outdoor gear, such as tents, hiking boots, and backpacks. She actively researches and reads reviews to ensure durability and functionality.&Daniel owns a pet store and constantly seeks out pet-related products for his business. He actively sources pet food, toys, grooming supplies, and accessories to cater to various pet owners' needs.&Olivia loves finding the best deals and discounts. She enjoys using coupons, comparing prices, and exploring online platforms to save money on her purchases. She values both quality and affordability.\\
\midrule
\textbf{Profile}&\textbf{Role \#11}&\textbf{Role \#12}&\textbf{Role \#13}&\textbf{Role \#14}&\textbf{Role \#15}\\
\midrule
Name&Ahmed&Sophia&Carlos&Emily&Javier\\
Gender&Male&Female&Male&Female&Male\\
Age&38&25&30&27&34\\
Occupation&Physical Education Teacher&Social Media Influencer&Automotive Engineer&Environmental Activist&Travel Blogger\\
Habits&Ahmed is passionate about sports and fitness. He shops for athletic apparel, sports equipment, and supplements. He enjoys exploring local sports stores and stays updated on the latest fitness trends.&Sophia is a beauty enthusiast and creates content about cosmetics, skincare, and makeup tutorials. She actively seeks out new beauty products, follows trends, and shares her recommendations with her followers.&Carlos has a deep interest in cars and enjoys shopping for automotive accessories, performance parts, and maintenance tools. He actively researches and stays updated on the latest automobile technology.&Emily is focused on sustainable living and seeks out eco-friendly products. She shops for ethically sourced clothing, reusable items, and environmentally friendly household products.&Javier loves traveling and exploring new destinations. He shops for travel gear, luggage, and outdoor accessories. He values lightweight and durable products for his adventures.\\
\bottomrule
\end{tabular}
\label{table:roles_part1}
\end{table*}

\begin{table*}[t]
\centering
\small
\caption{The generated 30 different simulated users by using ChatGPT (part 2).}
\begin{tabular}{lp{2.5cm}p{2.3cm}p{3cm}p{2.8cm}p{2.2cm}}
\toprule
\textbf{Profile}&\textbf{Role \#16}&\textbf{Role \#17}&\textbf{Role \#18}&\textbf{Role \#19}&\textbf{Role \#20}\\
\midrule
Name&Lily&Oliver&Emma&Noah&Ava\\
Gender&Female&Male&Female&Male&Female\\
Age&29&19&52&65&45\\
Occupation&Marketing Manager&College Student&Antique Store Owner&Retired Engineer&Art Gallery Owner\\
Habits&Lily recently became a new parent and actively shops for baby products, including clothing, toys, and nursery essentials. She seeks out trusted brands and prioritizes safety and quality.&Oliver is passionate about music and loves shopping for musical instruments, equipment, and vinyl records. He actively explores local music stores and online platforms for unique finds.&Emma has a keen interest in vintage items and actively seeks out antique furniture, clothing, and collectibles. She enjoys visiting flea markets, estate sales, and auctions to expand her collection.&Noah embraces technology and enjoys shopping for the latest gadgets, smartphones, and smart home devices. He actively seeks user-friendly products and keeps up with technological advancements.&Ava is passionate about art and actively collects paintings, sculptures, and other fine art pieces. She frequents art fairs, galleries, and auctions to discover new artists and expand her collection.\\
\midrule
\textbf{Profile}&\textbf{Role \#21}&\textbf{Role \#22}&\textbf{Role \#23}&\textbf{Role \#24}&\textbf{Role \#25}\\
\midrule
Name&Max&Olivia&Liam&Sophia&Ethan\\
Gender&Male&Female&Male&Female&Male\\
Age&20&35&32&26&50\\
Occupation&College Student&Interior Designer&Personal Trainer&Graphic Designer&Landscape Architect\\
Habits&Max is an avid gamer and actively shops for the latest gaming consoles, accessories, and video games. He stays updated on gaming news, follows esports tournaments, and seeks out merchandise from his favorite games.&Olivia specializes in creating beautiful spaces and frequently shops for furniture, decor, and lighting fixtures. She stays updated on design trends, visits trade shows, and sources unique pieces for her clients.&Liam is dedicated to fitness and actively shops for workout apparel, equipment, and supplements. He seeks out high-quality gear that withstands intense training sessions and recommends products to his clients.&Sophia has a passion for stationery and actively shops for notebooks, pens, art supplies, and planners. She values aesthetically pleasing and functional products that inspire her creativity.&Ethan enjoys gardening and frequently shops for plants, seeds, gardening tools, and outdoor decor. He seeks out sustainable and eco-friendly products that enhance his garden.\\
\midrule
\textbf{Profile}&\textbf{Role \#26}&\textbf{Role \#27}&\textbf{Role \#28}&\textbf{Role \#29}&\textbf{Role \#30}\\
\midrule
Name&Mia&Noah&Isabella&James&Harper\\
Gender&Female&Male&Female&Male&Female\\
Age&38&75&30&35&23\\
Occupation&CEO&Retired Teacher&Environmental Scientist&Stay-at-home dad&Vintage Clothing Store Owner\\
Habits&Mia appreciates luxury and actively shops for high-end fashion, accessories, and designer items. She seeks out exclusive brands, attends fashion events, and values premium craftsmanship.&Noah prefers simplicity when it comes to technology and shops for user-friendly devices, such as easy-to-use smartphones, tablets, and assistive technology. He values products with clear instructions and reliable customer support.&Isabella is committed to sustainable living and actively shops for eco-friendly products, including reusable bags, zero-waste toiletries, and environmentally friendly cleaning supplies. She values products with minimal environmental impact.&James is a hands-on parent and frequently shops for baby gear, including strollers, baby carriers, and childproofing items. He seeks out functional and safe products that make parenting easier.&Harper has a passion for vintage fashion and actively shops for retro clothing, accessories, and antique jewelry. She enjoys visiting thrift stores, vintage markets, and online platforms for unique finds.\\
\bottomrule
\end{tabular}
\label{table:roles_part2}
\end{table*}

\textbf{Human Evaluation for the Ranking Results by ChatGPT.} Furthermore, we conducted a human evaluation on the ranking results of ChatGPT, and the results in Table~\ref{table:ndcg} show that the NDCG score of ChatGPT is 0.871, indicating that the simulation results are close to human ranking preferences.

\clearpage

\begin{table*}[h!]
\centering
\small
\caption{NDCG@10 scores of the ranking results according to the results by ChatGPT and eight annotators.}
\begin{tabular}{ccccccc}
\toprule
    \textbf{Annotator}&\textbf{Query \#1}&\textbf{Query \#2}&\textbf{Query \#3}&\textbf{Query \#4}&\textbf{Query \#5}& \textbf{Average}.\\
    \midrule
\#1&0.807	&0.830	&0.826	&0.857	&0.819	&0.828\\
\#2&0.823	&0.908	&0.871	&0.798	&0.814	&0.843\\
\#3&0.795	&0.956	&0.999	&0.962	&0.964	&0.935\\
\#4&0.927	&0.896	&0.974	&0.931	&0.763	&0.898\\
\#5&0.843	&0.900	&0.910	&0.961	&0.876	&0.898\\
\#7&0.836	&0.934	&0.883	&0.860	&0.915	&0.885\\
\#6&0.851	&0.905	&0.872	&0.749	&0.841	&0.844\\
\#8&0.798	&0.824	&0.920	&0.764	&0.877	&0.837\\
\midrule
Average.&0.835	&0.894	&0.	907&0.	860&0.859	&0.871\\
\bottomrule
\end{tabular}
\label{table:ndcg}
\end{table*}

\subsection{Runtime Environment} \label{appendix-1-3}

To measure the expenses of the agents themselves during the operating process, we implement the runtime environment for the agent working system that tracks the temporal and monetary expenditures. We compute the monetary cost of each API call of the proprietary foundation models based on their official pricing. We also track the time consumption of the interaction between the agents and the environment. As the response of the website can be affected by networking issues, we leverage the following benchmarking measures for simplicity: In our environment construction, we documented all kinds of actions taking place in the environment and pre-calculated a static list of their estimated response delays. We also disregard the duration of API calls in the runtime environment as the tracked monetary cost also reflects the expense of API calls. With the runtime environment as a wrapper of the working system, human users can monitor the obedience of the agents to their self-constraints.

Specifically, we estimate the response delay of each action in the interactive environment by artificially sampling actions, trying them out, and then recording the delays. After gathering the data of time delays for each action, we fit the data with a uniform distribution, and use the expected value to reflect the estimated time for the action. Note that this estimated time is static, and is leveraged to benchmark the time cost in the experiments. The estimated time delays for all the actions are listed in the Table~\ref{table:simutime}.

\begin{table*}[h!]
\centering
\small
\caption{Estimated time for different actions.}
\begin{tabular}{|l|l|}
\toprule
    \textbf{Action} &\textbf{Time (s)} \\
    \midrule
    \texttt{reset} &0.1874 \\
    \verb|search| &0.5966 \\
    \verb|click[Instruction History]| &0.2645 \\
    \verb|click[Back to Search]| &0.1197 \\
    \verb|click[Next >]| &0.2693 \\
    \verb|click[< Prev]| &0.2545 \\
    \verb|click[Descriptions]| &0.2401 \\
    \verb|click[Features]| &0.2167 \\
    \verb|click[Reviews]| &0.1275 \\
    \verb|click[Buy Now]| &0.1920 \\
    \verb|click[other valid tag]| &0.2896 \\
    \verb|think| &0.0000 \\
    \midrule
    Invalid Action &0.3234 \\
    \midrule
    \textbf{Average} &0.2370 \\
\bottomrule
\end{tabular}
\label{table:simutime}
\end{table*}

\clearpage

\section{Experiment}\label{app:exp}

\begin{figure*}[t]
    \centering
    \includegraphics[width=\linewidth]{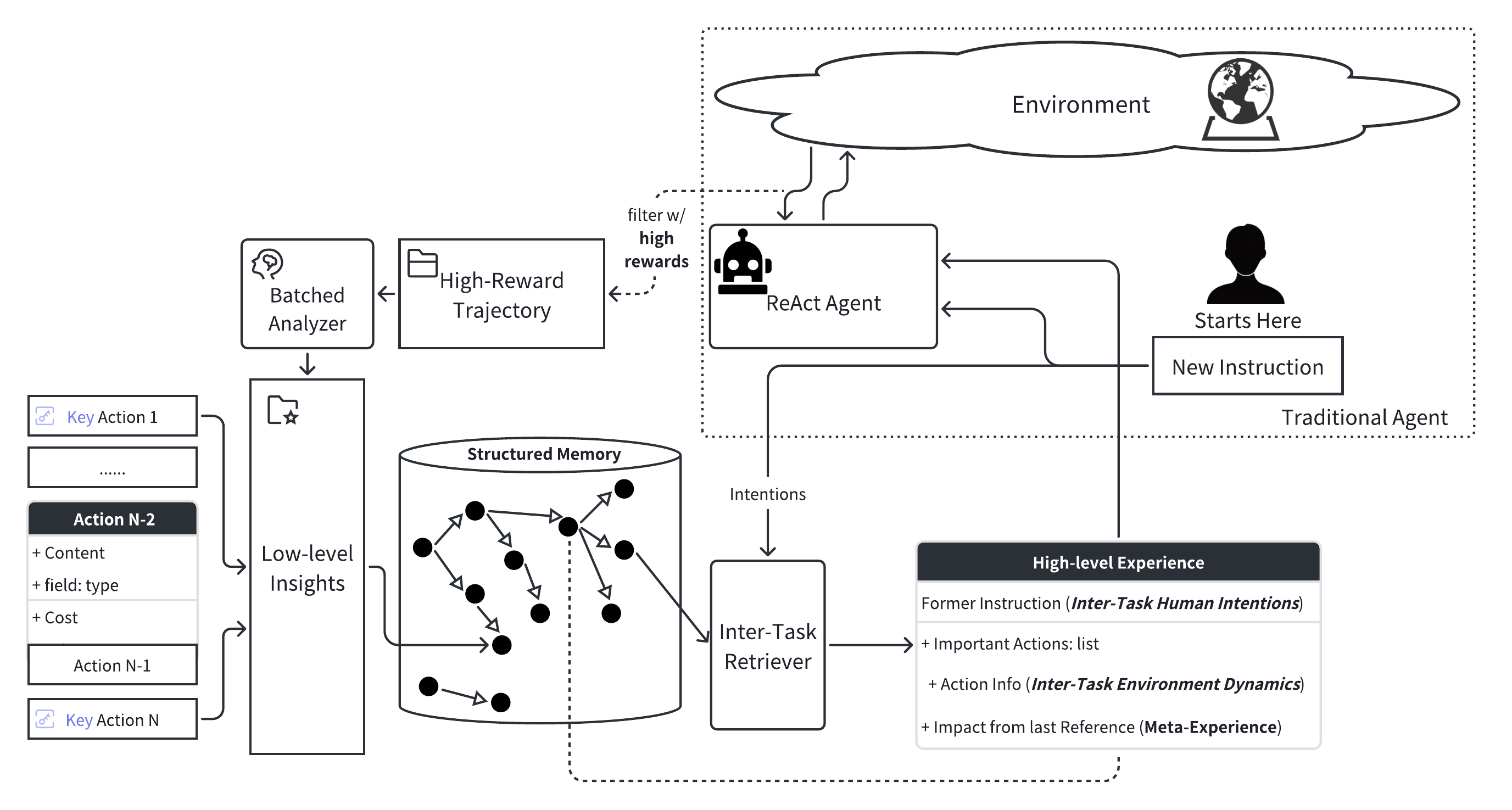}
    \caption{The details of our agent design that follows the principles of \textbf{UA}$^2$. Compared to traditional \texttt{ReACT} agents, we append structured experience as the long-term memory: By filtering and analyzing raw trajectories, we extracted key actions from prior successes as low-level insights in reasoning/action paths. By retrieving reference low-level insights under the same user, we can find the high-level experience under most similar user instructions, expressing similar human intentions. Agents are able to understand human intentions and environment dynamics by extrapolating key actions from a similar, prior task.}\label{fig:app-method}
\end{figure*}

\subsection{Descriptions on Agent Design with the Principles of UA$^2$}

From \cref{sec-32}, we identify three core capacities essential to agents and how they are related to the alignment principles. However, the trade-off between stronger capacities and fulfilling these principles makes it challenging to design a unified method that achieves both. Instead, to make the very first step, we considered expanding simple techniques that satisfy most principles to augment various capacities.

We introduce structured experience for unified Alignment principles for Agents (\textbf{UA}$^2$), as depicted in \cref{fig:app-method}, in addition to original ReAct~\citep{yao2022react} agent. 
\begin{definition}\label{def:traj}
    Trajectory is a list of actions ($a_i$) and observations ($o_i$) that agents have observed from environment ($\mathcal{E}$) after each action: $\mathcal{T} = \{a_i, o_i\}_{i=1}^{n}, o_i = \mathcal{E}(\{a_k\}_{k=1}^{i})$.
\end{definition}
After filtering high-reward trajectories (\cref{def:traj}) under human intentions from the given instruction $q$ and temporal environment conditions ($\mathcal{E}$), we utilize a batched analyzer~\citep{cheng-etal-2023-batch} that tags key actions from the whole trajectory within one API call. Thus we integrate insights from environment dynamics with low costs compared to Reflexion~\citep{shinn2023reflexion} and LATS~\citep{zhou2023language}.
\begin{definition}\label{def:keyActions}
    Key actions ($a_i^*$) are those having a positive impact on the efficiency or efficacy of task completion. We obtain the key actions $\mathcal{T}^*_q = \{a^*_i\}_{i=1}^{m}$ when the agent completes a task each time.
\end{definition}
\color{black}
\begin{definition}\label{def:shortcut}
    Low-level insights is a list of key actions ($a^*_i$) under a given instruction ($q$): $\mathcal{T}^*_q = \{a^*_i\}_{i=1}^{m}$. On top of this, structured experiences at $t$-th query are formed by a set of paired previous instructions and key actions: $\mathcal{S} = \{(q, \mathcal{T}^*_q): q \in \mathcal{Q}_{t-1}\}$, where $\mathcal{Q}_{t-1}$ denotes the set of previous instructions before $t$-th query.
\end{definition}
Thus, we enhance agents' memory by mapping low-level insights with corresponding instructions of previous tasks, which formulates a structured experience ($\mathcal{S}$)~(\cref{def:shortcut}). It is worth noting that we construct the structured memory under each user, which allows agents to comprehend human intention from prior instructions. When a new instruction is given, denoted as $q_{given}$, a reference contains high-level experience under the instruction $q_r$ could be retrieved by simply calculating the most similar instructions from its memory using \textit{BM25}~\citep{robertson2009probabilistic} scoring: $\mathcal{T}^*_{q_r} \in \mathcal{S}, q_r = \mathrm{argmax}_{q\in \mathcal{Q}}\{\mathrm{BM25}(q, q_{given})\}  $. 

The retrieved high-level experience acts as a plan across tasks which guides the agent towards the goal accurately and rapidly, also reducing costs of time and money. After completing a task under the environment, the agent analyze the new trajectory as well as the experience learnt from the former reference, termed ``meta-experience''. We then record the the new insights adjacent to the former reference. Also, we copy the ``meta-experience'' in attach to the former reference, which could be retrieved for in-coming tasks.

\subsection{Implementation Details}
We evaluate our method and baseline methods across all 10 users on our retrofitted Webshop, each comprising 50 tasks, except for LATS which is evaluated with only one user due to its high cost. All methods utilize \texttt{gpt-3.5-turbo-instruct-0914} as the underlying model for their agents except for LATS where we utilize \texttt{gpt-3.5-turbo-1106} to keep the same setting as the original paper. In executing each task, we limited the interaction with the web to a maximum of 15 steps per task, inclusive of any invalid actions.  

\subsection{Details of Experimental Setups}
For ReAct, Reflexion, and our method, we set the temperature as \textrm{0.0}. For ReAct-SC, we set the number of samples $k$ to be \textrm{3} and the temperature to be \textrm{0.05}. We also experiment with the family of chain-of-thought methods: CoT~\cite{wei2022chain}, CoT-L2M~\cite{zhou2022least}, CoT-SC~\cite{wang2022self}. Since their performances in task completion are significantly less competitive than other methods (see the next section), we exclude them from the main experiments in \cref{sec-43}, but include them here for reference. For CoT and CoT-L2M, we also set the temperature as \textrm{0.0}; and for CoT-SC, we also set  $k=3$ and the temperature to be \textrm{0.05}. To adhere to the same settings with \cite{zhou2023language}, we set the temperature to be 1.0, $k$ to be \textrm{5}, the number of iterations $n$ to be \textrm{30} for LATS.

All methods were tested in each of the following three environments respectively:
\begin{itemize}
    \item The fully retrofitted environment: configured exacted as described in \cref{sec-41} and \cref{appendix-1}.
    \item The ablated environment that excludes \textit{human intentions}: Based on the fully retrofitted environment, the hidden attributes from the user profiles and to be inferred from purchase histories are not excluded from reward computation. The alignment gap with \textit{human intentions} ($\mathbf{G}_{\mathrm{HI}}$) can be identified by comparing the test performances therein with those in the fully retrofitted environment.
    \item The ablated environment that excludes \textit{environmental dynamics}: The fully retrofitted environment with the re-ranking algorithms in \cref{appendix-1-2} disabled. The alignment gap with \textit{environmental dynamics} ($\mathbf{G}_{\mathrm{ED}}$) can be identified by comparing the test performances therein with those in the fully retrofitted environment.    

\end{itemize}

\subsection{Results}
\begin{figure}[h]
 \begin{minipage}{0.48\linewidth}
 	\vspace{3pt} 	\centerline{\includegraphics[width=\textwidth]{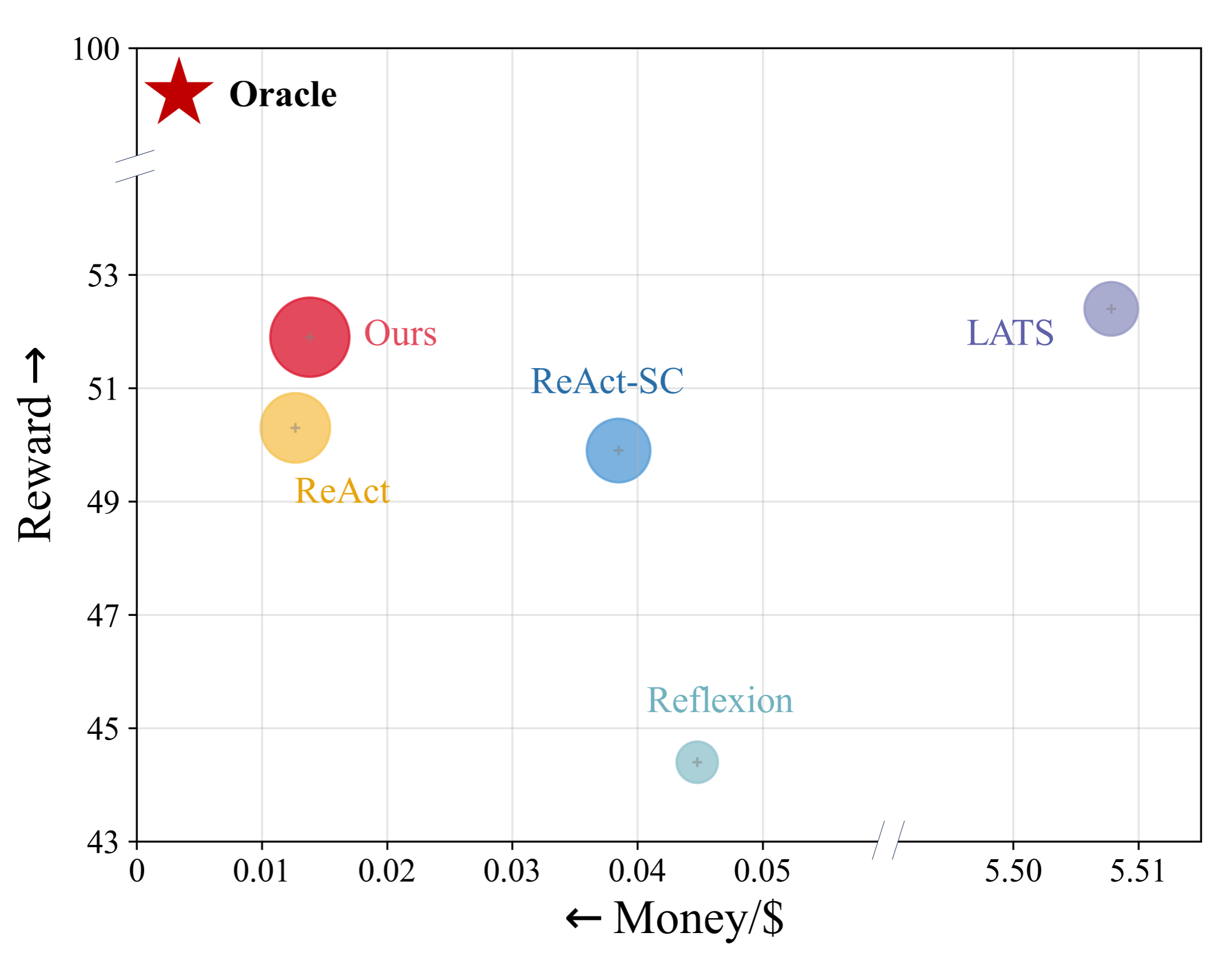}}
 
 \end{minipage}
 \begin{minipage}{0.48\linewidth}
	\vspace{3pt}	\centerline{\includegraphics[width=\textwidth]{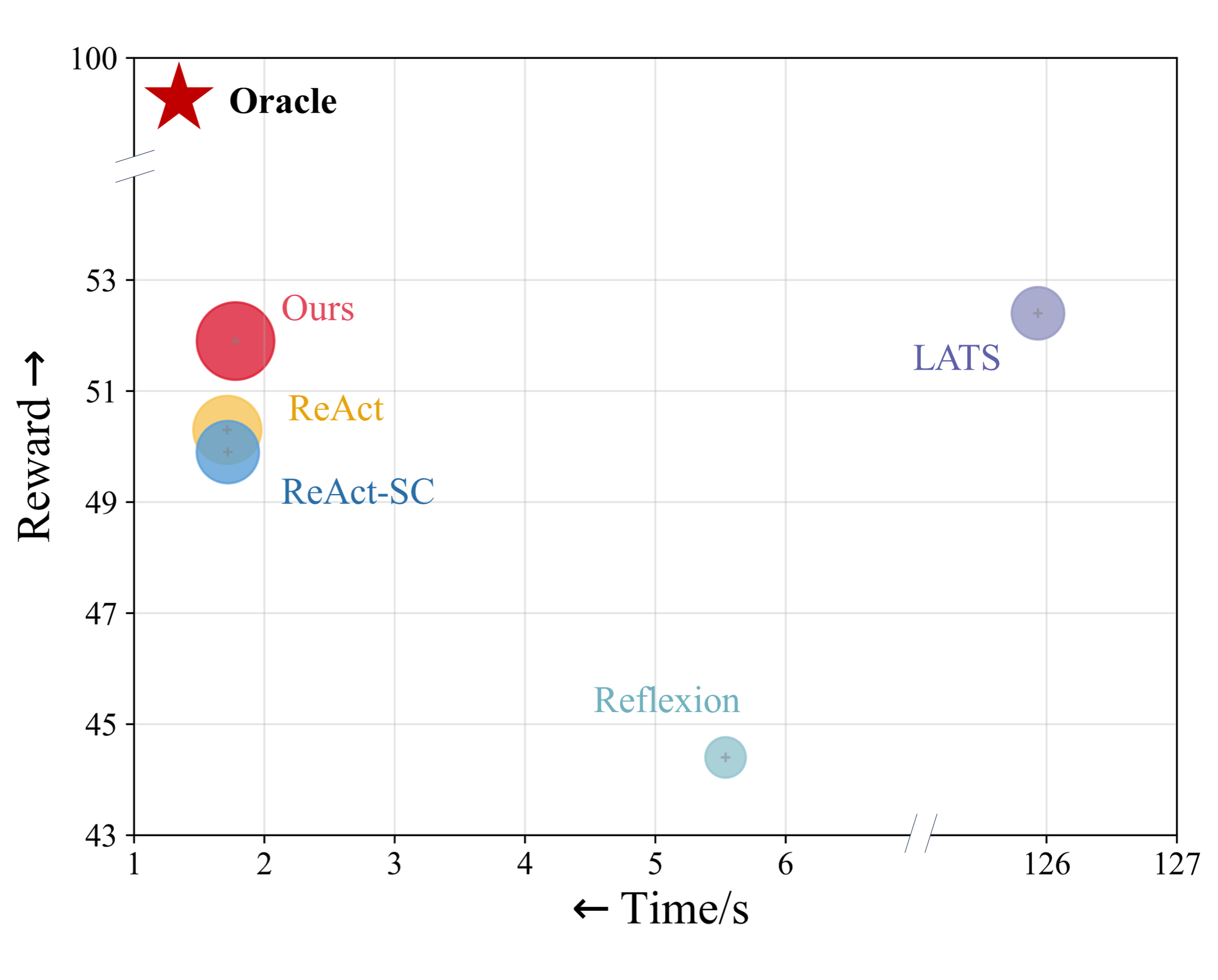}}

\end{minipage}
\caption{Agent's performance against the alignment gap with self-constraints tested on the retrofitted WebShop. The size of each circle represents the alignment gap with \textit{human intentions} ($\mathbf{G}_{\mathrm{HI}}$). The red star symbolizes our ultimate goal of developing an oracle agent capable of flawlessly completing complex tasks with minimal cost.}
\label{fig:mainRes4}
\end{figure}

\begin{figure}[h]
 \begin{minipage}{0.48\linewidth}
 	\vspace{3pt} 	\centerline{\includegraphics[width=\textwidth]{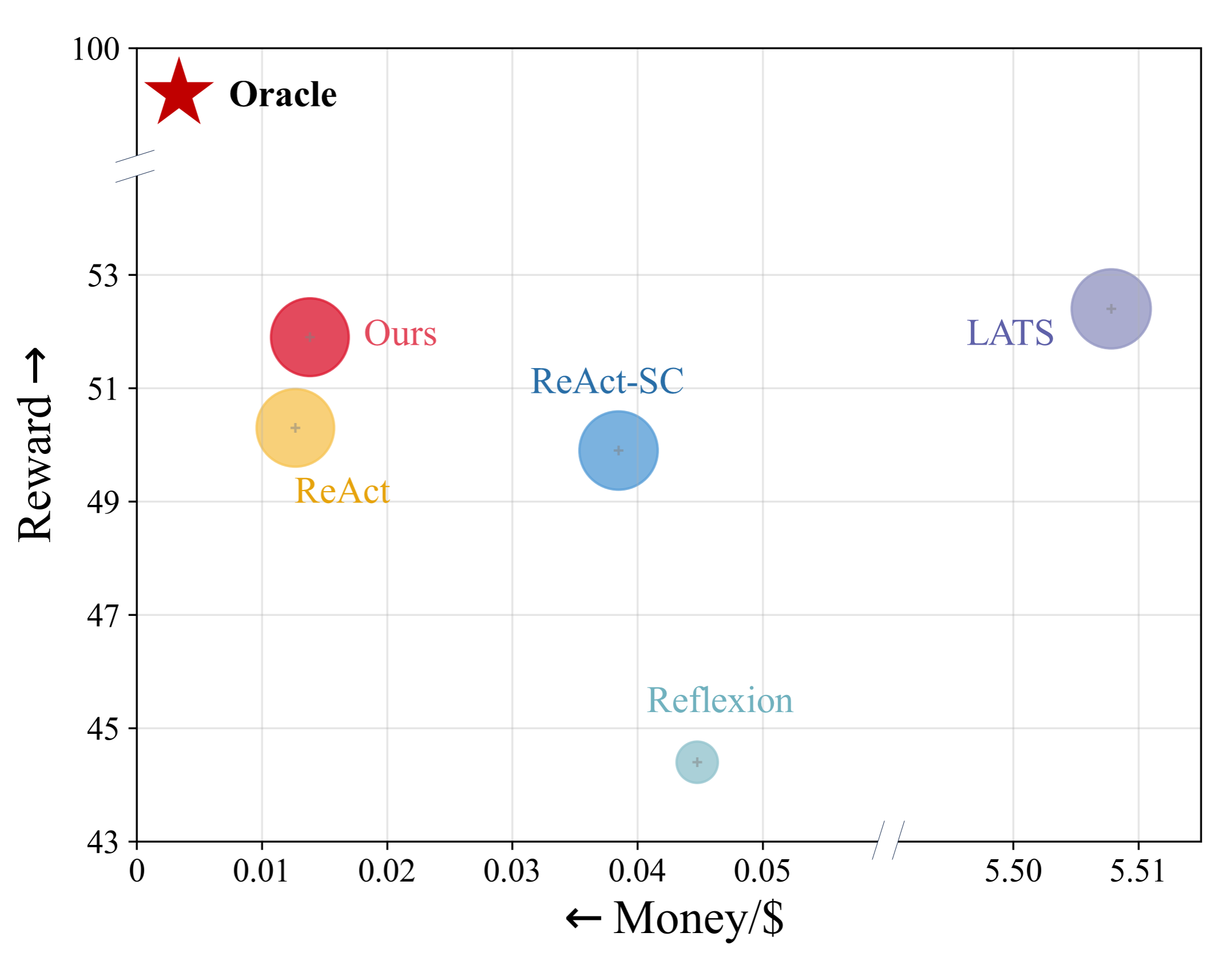}}
 
 \end{minipage}
 \begin{minipage}{0.48\linewidth}
	\vspace{3pt}	\centerline{\includegraphics[width=\textwidth]{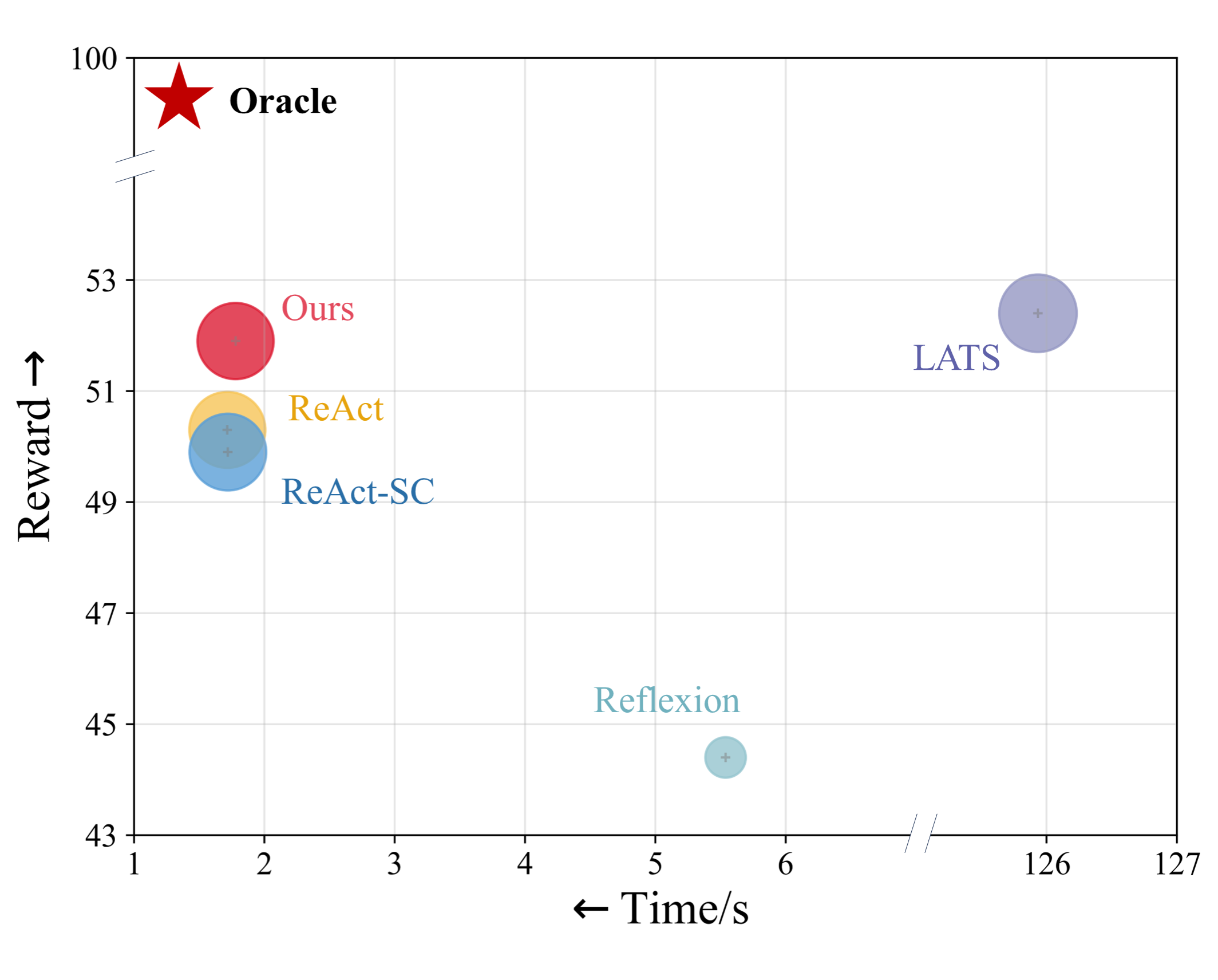}}

\end{minipage}
\caption{Agent's performance against the alignment gap with self-constraints tested on the retrofitted WebShop. The size of each circle represents the alignment gap with \textit{environmental dynamics} ($\mathbf{G}_{\mathrm{ED}}$). The red star symbolizes our ultimate goal of developing an oracle agent capable of flawlessly completing complex tasks with minimal cost.}
\label{fig:mainRes8}
\end{figure}

We present the comparative results on our retrofitted WebShop in Figure \ref{fig:mainRes4}, Figure \ref{fig:mainRes8}, and Table \ref{tab:overallRes}. 
Note that due to the significantly lower reward and success rate of CoT-related methods compared to others, the relative differences $\mathbf{G}_{\mathrm{HI}}$ and $\mathbf{G}_{\mathrm{ED}}$ can be dominantly affected by stochastic issues, and are therefore not of reference and comparison value.

In each figure, the X-axis represents the alignment gap with \textit{self-constraints}, and the Y-axis represents the performance. In terms of success rate, our proposed agent demonstrates comparable performance to Reflexion. To be specific, our approach places a greater emphasis on minimizing costs, whereas Reflexion prioritizes performance improvement. Our proposed agent, guided by the principles of \textbf{UA}$^2$, achieves a good balance between reward and cost considerations, while there remain a substantial gap between our agent and the ultimate goal of an oracle agent. 

\begin{table*}[t]
    \caption{Reward, success rate (SR), alignment gap with human intentions and environment dynamics, time and money cost result on our retrofitted WebShop.  *LATS is tested on 1/10 subset of the entire task collection due to the significant cost.}
    \label{tab:overallRes}
    \centering
    \begin{tabular}{l|cc|cccc}
    \toprule
    Method & Reward $\uparrow$ & SR (\%) $\uparrow$ & $\mathbf{G}_{\mathrm{HI}}$ (\%) $\downarrow$ & $\mathbf{G}_{\mathrm{ED}}$ (\%) $\downarrow$ & Time (s) $\downarrow$ & Money (\$) $\downarrow$ \\
    \midrule
    CoT          & ~~8.5& ~~1.2 & 22.4 & 67.1 & ~~~~1.858   & 0.011 \\
    CoT-L2M    & ~~9.8  & ~~0.8 & ~~6.1  & ~~8.2  & ~~~~1.939 & 0.037 \\
    CoT-SC     & 11.8 &   ~~1.4 & 32.2 & ~-61.9~~ & ~~~~1.883  & 0.032 \\
    \midrule
    ReAct      & 50.3 & ~~8.0 & 11.7 & 14.9 & ~~~~1.716 & 0.013 \\
    ReAct-SC   & 49.9 & ~~7.4 & 14.4 & 14.6 & ~~~~1.720 & 0.039 \\
    Reflexion  & 44.4 & 13.8 & 22.5 & 25.7 & ~~~~5.539 & 0.045 \\
    LATS*       & 52.4 & 10.0 & 18.5 & 14.3 & 125.935 & 5.508 \\
    \midrule
    Ours       & 51.9 & ~~9.6 & ~~6.7 & 14.8 & ~~~~1.779 & 0.014 \\
    \bottomrule
    \end{tabular}
\end{table*}

\end{document}